%% file: neurips_2026.tex
\documentclass{article}

\usepackage{amsmath}
\usepackage{booktabs}
\usepackage{enumitem}
\usepackage{hyperref}
\usepackage[capitalize, noabbrev]{cleveref}
\usepackage{subcaption}

\usepackage{algorithm}
\usepackage{algpseudocode}
\usepackage{pifont}        

\usepackage{graphicx}
\usepackage{amsmath}
\usepackage{cleveref}
\usepackage{multirow}
\usepackage{linguex}
\usepackage{sidecap}
\usepackage[utf8]{inputenc}
\usepackage[T1]{fontenc}

\usepackage{amssymb}
\usepackage{amsthm}

\usepackage{amsthm, amsmath, amssymb}

 \usepackage[preprint]{neurips_2026}

\usepackage[utf8]{inputenc} 
\usepackage[T1]{fontenc}    
\usepackage{hyperref}       
\usepackage{url}            
\usepackage{booktabs}       
\usepackage{amsfonts}       
\usepackage{nicefrac}       
\usepackage{microtype}      
\usepackage{xcolor}         

\usepackage{wrapfig}
\usepackage[most]{tcolorbox}
\tcbuselibrary{breakable}
\definecolor{promptbg}{RGB}{245, 248, 250}
\definecolor{promptborder}{RGB}{76, 175, 147}
\definecolor{prompttitle}{RGB}{76, 175, 147}

\title{CAPS: Cascaded Adaptive Pairwise Selection for Efficient Parallel Reasoning}

\author{
  \textbf{Fangzhou Lin}$^{a,b,c}$ {\hspace{.1em}}\quad
  \textbf{Shuo Xing}$^{a}$ {\hspace{.1em}}\quad
  \textbf{Peiran Li}$^{a}$ {\hspace{.1em}}\quad
  \textbf{Siyuan Yang}$^{a}$ {\hspace{.1em}}\quad
  \textbf{Qianwen Ge}$^{d}$ {\hspace{.1em}}\quad
  \vspace{.5em}\\
  \textbf{Kazunori Yamada}$^{c}$ {\hspace{.1em}}\quad
  \textbf{Ziming Zhang}$^{b}$ {\hspace{.1em}}\quad
  \textbf{Haichong Zhang}$^{b}$ {\hspace{.1em}}\quad
  \textbf{Zhengzhong Tu}$^{a}$ {\hspace{.1em}}\quad
  \vspace{.5em}\\
  $^a$Texas A\&M University \quad $^b$Worcester Polytechnic Institute \\
  $^c$Tohoku University \quad $^d$Georgia Institute of Technology
}

\begin{document}

\maketitle

\input{sec/abstract}

\input{sec/intro}


\input{sec/relatedwork}

\input{sec/method}
\input{sec/experiment}

\input{sec/conclusion}

\bibliographystyle{plain}
\bibliography{references}

\newpage
\appendix

\input{sec/supply_relatedwork}

\input{sec/X_suppl}

\input{sec/exp_suppl}

\input{sec/exp_imp_suppl}




\end{document}

%% file: sec/abstract.tex
\begin{abstract}
Parallel reasoning, where a generator samples many candidate solutions and an aggregator selects the best, is one of the most effective forms of test-time scaling in large language models, and pairwise self-verification has become its strongest aggregation primitive. Yet pairwise verification carries a heavy cost: each judgment reads two complete solutions in full, and existing methods perform tens of such judgments per problem regardless of whether the comparison is informative. We introduce \textbf{CAPS} (Cascaded Adaptive Pairwise Selection), an inference-only framework that allocates verifier compute non-uniformly along two orthogonal axes: an \emph{evidence axis} that adapts how much of each candidate the judge sees, and a \emph{distribution axis} that adapts how comparisons are spread across the pool. CAPS instantiates these into a four-stage cascade with an optional rescue subroutine, and admits a closed-form verifier-token cost in which the per-candidate marginal cost is roughly halved relative to uniform full-evidence schedules. On four self-verifying models (Qwen3-14B, GPT-OSS-20B, Qwen3-4B-Instruct/Thinking) and five reasoning benchmarks spanning code (LiveCodeBench-v5/v6, CodeContests) and math (AIME 2025, HMMT 2025), CAPS outperforms the leading pairwise verifier on $14$ of $20$ suites while using $25.4\%$ of its verifier-token budget on code, and outperforms pointwise self-verification on \emph{all} $20$. The trade-off suites admit an interpretable diagnostic in terms of the verifier's accuracy at partial versus full evidence, providing a concrete pre-deployment check for cascade suitability. 

\end{abstract}

%% file: sec/intro.tex
\section{Introduction}
\label{sec:intro}

Large language models have made substantial progress on reasoning tasks through inference-time scaling, where extra compute is spent to refine, reflect on, or diversify candidate solutions~\citep{snell2024scaling, wei2022chain, wang2022self, kojima2022large, snell2025scaling}. This compute can be deployed sequentially: a single chain of thought made longer through reflection or revision; or in parallel, by sampling many independent chains and aggregating their answers~\citep{lightman2023let,jaech2024openai,chow2024inference}. The latter also called \emph{Parallel reasoning}~\citep{brown2024large, snell2024scaling, singh2026v_1}, which has emerged as one of the most effective forms of test-time scaling: compared with single answer, the probability of correct answer increased significantly even at parallel number $N$ . As $N$ grows, the limiting factor for parallel reasoning shifts from the diversity of candidates to the reliability of the aggregator: parallel reasoning is only as good as the mechanism that picks the correct candidate from the pool~\citep{lightman2023let, jaech2024openai, chow2024inference, brown2024large, snell2024scaling, singh2026v_1}.

Self-verification, where the same model judges its own candidates during (parallel) reasoning, has become the dominant aggregator in domains with verifiable rewards such as code and math~\citep{weng2023large, huang2023large, stechly2024self}. The simplest form is \emph{pointwise scoring}~\cite{venkatraman2025recursive}: prompting the model to assign an absolute quality rating to each candidate in isolation. This approach suffers from calibration failures: ratings saturate at the top of the scale and lose discriminative power~\citep{weng2023large, zhuang2024setwise, stechly2024self}; the verifier exhibits self-preference bias, positively rating its own samples even when they are incorrect~\citep{madaan2025rethinking}; and ratings produced for different candidates lack a globally comparable scale, because each judgment is made without reference to any other candidate~\citep{singh2026v_1}. Recent work has converged on \emph{pairwise comparison} as a more robust primitive: head-to-head judgments only require the verifier to determine in which of two solutions is better: a relative judgment that sidesteps the calibration issues of absolute scoring. The leading method in this regime, V1-Infer~\citep{singh2026v_1}, uses an uncertainty-guided Swiss tournament~\citep{csato2017ranking} with roughly $48$ pairwise comparisons, demonstrating substantial gains over pointwise verification on code and math benchmarks.

Pairwise verification~\citep{stechly2024self, madaan2025rethinking, singh2026v_1, venkatraman2025recursive}, however, has its own bottleneck: heavy cost. Each judgment reads two complete $4$--$8$K-token solutions, so the verifier-token budget routinely exceeds the cost of generation itself, and the cost grows quadratically with the number of comparisons. While existing methods carefully optimize \textit{which} pairs to compare: through Swiss-system tournaments~\cite{csato2017ranking}, uncertainty-guided pair selection~\citep{singh2026v_1}, and Bradley-Terry-inspired schedules~\cite{hunter2004mm}. They treat \textit{how} each comparison is performed as a fixed quantity: every pair is evaluated by reading both solutions in full, regardless of whether the comparison decides between two strong contenders or two candidates that are clearly wrong. This uniformity is at odds with the structure of the selection problem itself. Most candidate pairs are easy to discriminate: they may produce different final answers, pursue visibly different algorithmic strategies, or differ in early reasoning steps that already determine correctness. And most candidates are not the eventual winner; the selection argmax depends only on the ordering among the strongest few. The natural question is then: \textit{can we exploit this structure to spend verifier compute more efficiently, without sacrificing selection quality?}

\begin{figure}[t]
    \centering
    \includegraphics[width=0.8\textwidth]{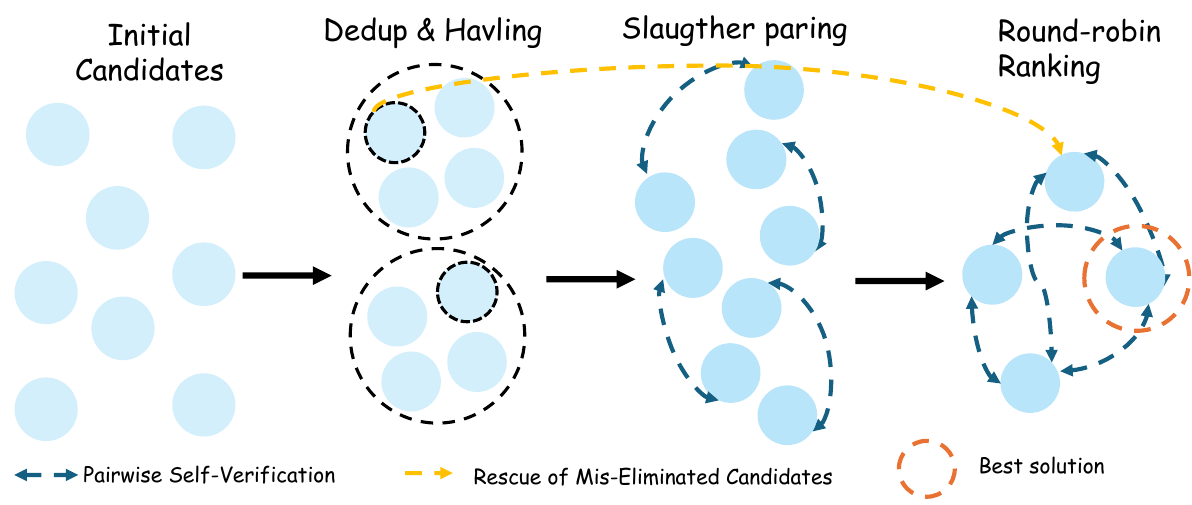}
    \caption{\textbf{CAPS Overview}. Deduplicate, eliminate at partial evidence, eliminate at full evidence, and round-robin among the finalists; with an optional rescue subroutine for cheap-evidence errors.}
    \label{fig:swiss_refinement_overview}
\vspace{-0.2cm}    
\end{figure}

To this end, we propose \textbf{\textit{CAPS}} (Cascaded Adaptive Pairwise Selection), an inference-only framework that allocates verifier compute non-uniformly along two orthogonal axes. The \emph{evidence axis} controls how much of each candidate the judge sees; we observe that most pairs admit a confident judgment from a partial view: the boxed answer, the first few lines of code, or a short reasoning window; at roughly an order of magnitude lower cost than a full read. The \emph{distribution axis} controls how comparisons are spread across the candidate pool; since the final selection depends only on the ordering among the strongest candidates, full-evidence comparisons should be reserved for them. CAPS instantiates these into a four-stage cascade with a closed-form verifier-token cost characterization, together with an optional rescue subroutine that protects against errors at the cheapest evidence level.

We evaluate CAPS against three baselines: Vanilla (no verification), Pointwise self-verification~\citep{venkatraman2025recursive}, and V1-Infer~\citep{singh2026v_1}; across four self-verifying models (Qwen3-14B~\citep{yang2025qwen3}, GPT-OSS-20B~\citep{agarwal2025gpt}, Qwen3-4B-Instruct-2507, and Qwen3-4B-Thinking-2507) and five reasoning benchmarks spanning code (LiveCodeBench-v5~\citep{jain2024livecodebench}, LiveCodeBench-v6, CodeContests\citep{li2022competition}) and math (AIME 2025, HMMT February 2025~\citep{balunovic2025matharena}). CAPS outperforms V1-Infer on $14$ of $20$ settings while using on average \textit{$25.4\%$} of its verifier-token budget on code, and outperforms Pointwise verification on \textit{all} $20$. The empirical token ratio matches our closed-form prediction to within three percentage points, indicating that the efficiency is structural rather than the result of hyperparameter tuning. The six results where V1-Infer is higher trail by at most $3.4$ Pass@1 points; we trace this trade-off to a measurable property of the deployment: the verifier's per-pair accuracy at partial vs.\ full evidence; giving practitioners a concrete pre-deployment check for cascade suitability.

\textbf{Contributions.}
Our contributions are summarized as follows.

\textbf{1.\ Two axes for verifier-compute allocation in pairwise selection.}
We identify the evidence level and the comparison distribution as orthogonal, underused degrees of freedom in pairwise self-verification. The evidence axis exploits the fact that most pairs are easy to discriminate at low evidence; the distribution axis exploits the fact that the selection argmax depends only on the strongest candidates. Together these reframe pairwise verification cost as a structural quantity.

\textbf{2.\ CAPS, a cascaded selection framework with closed-form cost.}
CAPS instantiates the two axes as a four-stage cascade: deduplicate, eliminate at partial evidence, eliminate at full evidence, and round-robin among the finalists; with an optional rescue subroutine for cheap-evidence errors.

\textbf{3.\ Empirical results spanning four models, five benchmarks, and four methods.}
Across our $20$ evaluation suites, CAPS outperforms V1-Infer in $14$ at roughly one-quarter of its verifier-token cost, and Pointwise verification in \textit{all} $20$. The empirical token ratio matches the analytical prediction to within three percentage points, and the trade-off cells admit an interpretable diagnostic in terms of the verifier's per-pair accuracy at partial vs.\ full evidence.

%% file: sec/relatedwork.tex
\section{Related Work}
\label{sec:related}

Inference-time scaling encompasses both \emph{sequential} approaches that lengthen a single chain of thought through reflection or revision~\citep{zhang2024chain, madaan2023self, qu2024recursive} and \emph{parallel} approaches that sample multiple independent chains and aggregate their answers~\citep{wei2022chain, wang2022self, brown2024large, snell2024scaling}; Self-Consistency~\citep{chen2023universal, wang2022self} and Best-of-$N$ sampling with learned reward models~\citep{lightman2023let, cobbe2021training} are the canonical aggregation primitives, and the accuracy gap on hard problems motivates stronger selection mechanisms~\citep{brown2024large, singh2026v_1}. Among these, \emph{self-verification}, which reuses the generator as a judge over its own samples~\citep{weng2023large, huang2023large, stechly2024self}, has become the dominant primitive in domains with verifiable rewards: pointwise scoring with each candidate rated in isolation suffers from documented calibration failures including score saturation, self-preference bias, and the lack of a globally comparable scale~\citep{stechly2024self, madaan2025rethinking, panickssery2024llm, zheng2023judging, venkatraman2025recursive}, and recent work has therefore converged on \emph{pairwise} comparison: V1-Infer~\citep{singh2026v_1} uses an uncertainty-guided Swiss tournament~\cite{csato2017ranking} to allocate a fixed comparison budget, and Pairwise RM~\citep{liu2025pairwise} trains a dedicated pairwise reward model with a knockout-tournament selection rule. An orthogonal line of work combines candidates rather than selecting among them, ranging from majority voting~\citep{chen2023universal} to learned aggregators such as AggLM~\citep{zhao2025majority} and Recursive Self-Aggregation~\citep{venkatraman2025recursive}; aggregation methods improve Pass@1 but exhibit diversity collapse, motivating hybrid designs that pair aggregation with verification~\citep{singh2026v_1}. A more detailed discussion of each of these threads is provided in Appendix~\ref{app:related-extended}.

%% file: sec/method.tex
\section{Method}
\label{sec:method}

We formulate selection from parallel reasoning candidates as a budgeted pairwise verification problem (\S\ref{sec:setup}), describe the four-stage CAPS pipeline that operationalizes it (\S\ref{sec:pipeline}--\S\ref{sec:rescue}), and analyze its verifier-token cost in closed form (\S\ref{sec:budget}). 
\vspace{-0.3cm}
\subsection{Problem Setup}
\label{sec:setup}
\vspace{-0.2cm}
\textbf{The selection task.}
Given a problem $q$ and $N$ candidate solutions $\mathcal{C} = \{c_1, \ldots, c_N\}$ sampled from a generator (any LLM), the selection task is to return an index 
\begin{equation}
    \hat{\imath} \;\in\; \arg\max_{i \in [N]} \,\Pr[\,y_i = 1 \mid q, \mathcal{C}\,],
    \label{eq:obj}
\end{equation}

where $y_i \in \{0, 1\}$ is the unobserved correctness of $c_i$. The verifier (same LLM in self-verification setting) accesses only a pairwise oracle and is constrained by a verifier-token budget $B$~\cite{venkatraman2025recursive,singh2026v_1}.

\textbf{Evidence map.}
Rather than treating each candidate as a monolithic object, we expose multiple views of it through an \emph{evidence map} $\{\phi_e\}_{e \in \mathcal{E}}$, indexed by detail level. Each level has a per-pair token cost $T_e$ monotone in $e$. We use three levels:
\begin{itemize}[leftmargin=1em, itemsep=0.5pt, topsep=1pt]
    \item $\phi_0(c) \in \Sigma$: discrete signature (boxed answer for math, AST hash for code), used for clustering only.
    \item $\phi_1(c) \in \mathcal{V}_1$: partial view (first ${\sim}20$ code lines, or boxed answer plus a ${\sim}300$-token reasoning window).
    \item $\phi_2(c) = c$: full solution.
\end{itemize}
We write $\rho := T_1/T_2 \in (0, 1)$ for the evidence-cost ratio, the governing quantity of the cost analysis in \S\ref{sec:budget}.

\textbf{Pairwise oracle.}
A pairwise judge at evidence level $e$ is a (possibly stochastic) map
\begin{equation}
    J\bigl(q,\, \phi_e(c_i),\, \phi_e(c_j)\bigr) \;\longmapsto\; (v_{ij},\, w_{ij}),
    \label{eq:judge}
\end{equation}
returning an outcome $v_{ij} \in \{0, \tfrac{1}{2}, 1\}$ from $i$'s perspective (loss, tie, win) and a confidence $w_{ij} \in [\tau_w, 1]$ floored at $\tau_w = 0.05$. Throughout the pipeline we maintain a per-candidate cumulative score $S(c)$, initialized to $0$ and updated after each judge call by
\begin{equation}
    S(c_i) \;\mathrel{+}=\; w_{ij}\, v_{ij}, \qquad
    S(c_j) \;\mathrel{+}=\; w_{ij}\, (1 - v_{ij}).
    \label{eq:score-update}
\end{equation}
Eq.~\eqref{eq:score-update} accumulates confidence-weighted outcomes: high-confidence wins dominate the score while ambiguous comparisons contribute little, so noisy judgments do not propagate into later decisions.

\textbf{Intuition.}
Two qualitative observations guide the design that follows. First, the argmax in Eq.~\eqref{eq:obj} depends only on the relative ordering among the strongest candidates: comparisons among low-quality candidates carry little information about the final selection. Second, many pairs admit a confident judgment from a partial view: candidates that differ in their final answer or in their opening algorithmic strategy can often be discriminated without reading either solution end-to-end. Together these suggest a verifier that reserves expensive full-evidence comparisons for a small subset of plausible winners and resolves the rest cheaply. They motivate the cascade structure but the empirical claims of the paper rest on \S\ref{sec:experiments}.

\subsection{The CAPS Pipeline}
\label{sec:pipeline}

CAPS proceeds in four stages, illustrated in Figure~\ref{fig:swiss_refinement_overview}: an equivalence-class deduplication (Stage~0), two halving rounds at increasing evidence (Stages~A and~B), and a complete round-robin among $f$ finalists at full evidence (Stage~C). Algorithm~\ref{alg:caps} gives the high-level procedure; the helpers \textsc{Dedup}, \textsc{Eliminate}, and \textsc{Rescue} are specified in Appendix~\ref{app:method}.


\begin{algorithm}[t]
\caption{CAPS — Cascaded Adaptive Pairwise Selection}
\label{alg:caps}
\begin{algorithmic}[1]
\Require problem $q$, candidates $\mathcal{C}$, judge $J$, finalist count $f$, weight floor $\tau_w$, rescue margin $\delta$
\Ensure selected candidate $c^\star$
\State $S(c) \gets 0$ for all $c \in \mathcal{C}$ \Comment{cumulative scores}
\State $\mathcal{C}' \gets \textsc{Dedup}(\mathcal{C}, \phi_0)$;\quad record cluster sizes $\nu(\cdot)$ \Comment{Stage 0 (no judge calls)}
\State $\mathcal{C}_A \gets \textsc{Eliminate}(\mathcal{C}',\, J,\, e{=}1,\, \text{seed by } \nu)$ \Comment{Stage A}
\State $\mathcal{C}_B \gets \textsc{Eliminate}(\mathcal{C}_A,\, J,\, e{=}2,\, \text{seed by } S,\, \text{stop at } f)$ \Comment{Stage B}
\If{CAPS-R enabled}\;\; $\mathcal{C}_B \gets \textsc{Rescue}(\mathcal{C}_B,\, \mathcal{C}'\setminus\mathcal{C}_B,\, \delta)$ \Comment{Eq.~\eqref{eq:rescue}}
\EndIf
\For{each unordered pair $\{c_i, c_j\} \subset \mathcal{C}_B$} \Comment{Stage C round-robin}
    \State $(v_{ij}, w_{ij}) \gets J(q,\, \phi_2(c_i),\, \phi_2(c_j))$
\EndFor
\State \Return $c^\star \gets \arg\max_{c \in \mathcal{C}_B} s_C(c)$ via Eq.~\eqref{eq:stagec}
\end{algorithmic}
\end{algorithm}

\textbf{Stage 0: Equivalence-class deduplication.}
We partition $\mathcal{C}$ via the equivalence relation $c_i \sim c_j \iff \phi_0(c_i) = \phi_0(c_j)$ and keep one representative per cluster, producing $\mathcal{C}'$ of size $N' \leq N$. Cluster sizes $\nu(c)$ are recorded as metadata for tie-breaking but \emph{do not contribute to $S(c)$}, preserving one-solution-one-vote semantics: a popular-but-wrong cluster majority should not outvote a correct singleton. When $\phi_0$ is informative (math benchmarks with closed-form answers), $N' \ll N$; on free-form code, $\phi_0$ rarely distinguishes candidates and Stage~0 reduces to a no-operation.

\textbf{Stage A: Halving at E1.}
We pair the $N'$ survivors using \emph{slaughter pairing}~\cite{sziklai2022efficacy}: the $i$-th and $(N'{+}1{-}i)$-th seeds, sorted by $\nu$ descending; query the judge at evidence level E1, update $S$ via Eq.~\eqref{eq:score-update}, and retain the higher-scoring member of each pair. It maximizes the seed gap within each pair, making the cheap E1 judgment more reliable: large quality differentials are easier to discriminate from a partial view than small ones. Stage~A makes $\lfloor N'/2 \rfloor$ E1 calls, and its principal residual risk is mis-elimination at E1; this is the failure mode that the rescue mechanism (\S\ref{sec:rescue}) addresses.

\textbf{Stage B: Reduction to $f$ at E2.}
Stage B halves the pool repeatedly at full evidence until the finalist count $f$ is reached, with pairing seeded by the current cumulative score $S$ rather than $\nu$. For $|\mathcal{C}_A| \leq 2f$ (the regime $N \leq 4f$, which covers our default $N=16, f=4$), Stage~B is a single round of $|\mathcal{C}_A|/2 = N'/4$ E2 calls; for larger pools it iterates until $|\mathcal{C}_{\text{cur}}| = f$. Each round updates $S$, so Stage~B's $S$ values reflect the integrated signal from both E1 and E2 rounds.

\textbf{Stage C: Round-robin at E2.}
The $f$ finalists are the candidates among which the argmax most plausibly lies. Stage~C concentrates the budget here: every unordered pair in $\mathcal{C}_B$ is judged once at E2, totaling $\binom{f}{2}$ comparisons. Each finalist's Stage-C score is the confidence-normalized win rate
\begin{equation}
    s_C(c) \;=\; \frac{\sum_{c' \in \mathcal{C}_B \setminus \{c\}} w(c, c')\, v(c, c')}{\sum_{c' \in \mathcal{C}_B \setminus \{c\}} w(c, c')},
    \label{eq:stagec}
\end{equation}
computed solely from Stage-C calls; the pre-Stage-C $S$ enters only as a tiebreaker (followed by $\nu$, then random). The denominator normalizes by aggregate confidence so $s_C$ measures the proportion of confidence-weighted wins rather than their unnormalized total. We use round-robin~\citep{shreedhar1996efficient,furnkranz2002round} rather than a Swiss~\citep{csato2017ranking} or bracket~\citep{tonon2004use} because, on a small finalist pool ($f = 4$), $\binom{f}{2}$ is already affordable and round-robin gives a path-independent ranking signal: every finalist is compared against every other finalist at full evidence. We return $c^\star = \arg\max_{c \in \mathcal{C}_B} s_C(c)$.

\subsection{CAPS-R: Rescue of Mis-Eliminated Candidates}
\label{sec:rescue}

Stage~A operates at E1 and is the most error-prone step. CAPS-R is an optional subroutine that admits one previously eliminated candidate into the finalist round when its elimination evidence appears weak. Let $c^+ := \arg\max_{c \in \mathcal{C}' \setminus \mathcal{C}_B} S(c)$ be the strongest eliminated candidate and $c_{\min} := \arg\min_{c \in \mathcal{C}_B} S(c)$ the weakest finalist. CAPS-R triggers if either
\begin{equation}
    \underbrace{|S(c^+) - S(c_{\min})| \,\leq\, \delta}_{\text{margin: lost narrowly}}
    \quad\text{or}\quad
    \underbrace{\nu(c^+) = 1 \;\text{and}\; |S(c^+) - S(c_{\min})| \,\leq\, 2\delta}_{\text{rarity: correct singleton}},
    \label{eq:rescue}
\end{equation}
in which case $c^+$ joins $\mathcal{C}_B$ and the round-robin expands from $\binom{f}{2}$ to $\binom{f+1}{2}$ E2 calls (overhead $f \cdot T_2$). We use $\delta = 0.15$. The two conditions cover distinct failure modes: \emph{margin} handles statistical noise on close pairs, while \emph{rarity} handles correct singletons whose stylistic distinctiveness can mislead a low-evidence judge; the relaxed threshold $2\delta$ for the rarity condition reflects the higher prior of misranking under stylistic mismatch.

\subsection{Verifier-Token Cost Analysis}
\label{sec:budget}

Let $T_{\text{CAPS}}(N', f)$ denote the total verifier-token cost of the deterministic pipeline. By construction of the four stages,
\begin{equation}
    T_{\text{CAPS}}(N', f) \;=\; \underbrace{\lfloor N'/2 \rfloor\, T_1}_{\text{Stage A}}
    \;+\; \underbrace{\Bigl(\sum_{r=1}^{r_B} \lfloor N_r/2 \rfloor\Bigr)\, T_2}_{\text{Stage B}}
    \;+\; \underbrace{\binom{f}{2}\, T_2}_{\text{Stage C}},
    \label{eq:cost-general}
\end{equation}
where $r_B := \lceil \log_2(\lceil N'/2 \rceil / f) \rceil$ is the number of Stage-B halving rounds and $N_r$ the round-$r$ entrant count. Closing the Stage-B geometric sum by telescoping (Appendix~\ref{app:cost-asymp}):
\begin{equation}
    T_{\text{CAPS}}(N', f) \;=\; \frac{N'}{2}(T_1 + T_2) \,-\, f\,T_2 \,+\, \binom{f}{2}\, T_2 \,+\, O(\log N').
    \label{eq:cost-asymp}
\end{equation}
The dominant term is linear in $N'$ with slope $\tfrac{1}{2}(T_1 + T_2)$: \emph{the marginal cost of an additional candidate is half a Stage-A E1 call plus half a Stage-B E2 call}, rather than the full $T_2$ that a uniform full-evidence schedule would incur. This replacement of $T_2$ by $\tfrac{1}{2}(T_1 + T_2)$ in the per-candidate marginal cost is the structural source of CAPS's efficiency.

For the standard configuration $N = N' = 16, f = 4$ (the worst case for CAPS, where Stage~0 yields no savings), Eq.~\eqref{eq:cost-general} gives $r_B = 1$, $N_1 = 8$, and
\begin{equation}
    T_{\text{CAPS}}(16, 4) \;=\; 8\,T_1 + 4\,T_2 + 6\,T_2 \;=\; 8\,T_1 + 10\,T_2 \;=\; (10 + 8\rho)\, T_2.
    \label{eq:cost-standard}
\end{equation}
With empirically measured $\rho \in [0.10, 0.15]$ (Appendix~\ref{app:tcost}), this evaluates to roughly $11\,T_2$: the entire pipeline costs approximately $11$ full-evidence comparison-equivalents per problem. The cost is determined by $N$, $f$, and $\rho$, all of which are properties of the deployment rather than free parameters.

%% file: sec/experiment.tex
\section{Experiments}
\label{sec:experiments}

We evaluate CAPS against three baselines on five reasoning benchmarks across four self-verifying models. Our goal is to measure (i) selection quality of the cascaded pipeline relative to existing pairwise verifiers, (ii) the verifier-token cost reduction afforded by progressive evidence, and (iii) the contributions of individual CAPS components.

\subsection{Experimental Settings}
\label{sec:exp-setup}

\textbf{Models and Benchmarks.}
We evaluate four models that span two families and two scale regimes: \textbf{Qwen3-14B}~\citep{yang2025qwen3}, \textbf{GPT-OSS-20B}~\citep{agarwal2025gpt}, \textbf{Qwen3-4B-Instruct-2507}~\citep{yang2025qwen3}, and \textbf{Qwen3-4B-Thinking-2507}~\citep{yang2025qwen3}. For code generation we use \textbf{LiveCodeBench-v5}~\citep{jain2024livecodebench} (279 problems, 24.08--25.02), \textbf{LiveCodeBench-v6}~\citep{jain2024livecodebench} (131 problems, 25.02--25.05), and \textbf{CodeContests}~\citep{li2022competition} (165 problems). For math reasoning we use \textbf{AIME 2025} (30 problems) and \textbf{HMMT February 2025} (30 problems)~\citep{balunovic2025matharena}. We sample $N = 16$ candidates per problem with temperature $0.6$ for code and $1.0$ for math (top-$p = 0.95$ throughout); the same model serves as both generator and pairwise judge. Generation prompts, full hyperparameters, and dataset sources are reproduced in Appendix~\ref{app:implementation}.

\begin{wraptable}{r}{8.5cm}
\centering
\caption{\textbf{Main results: Pass@1 (\%).} Selection accuracy across four methods, four models, and five benchmarks at $N = 16$, totaling $20$ (model, benchmark) suites. \textbf{Bold}: best method per (model, benchmark) row. AIME 2025 and HMMT 2025 each contain $30$ problems, so a single problem flip corresponds to $3.3$ percentage points.}
\label{tab:main}
\tiny
\setlength{\tabcolsep}{6pt}
\begin{tabular}{ll cccc}
\toprule
\textbf{Model} & \textbf{Benchmark} & \textbf{Vanilla} & \textbf{Pointwise} & \textbf{V1-Infer} & \textbf{CAPS} \\
\midrule
\multirow{5}{*}{Qwen3-14B}        & LCB-v5        & $26.2$ & $29.3$ & $34.8$ & $\mathbf{37.6}$ \\
                                  & LCB-v6        & $50.7$ & $53.2$ & $58.8$ & $\mathbf{59.5}$ \\
                                  & CodeContests  & $21.7$ & $24.5$ & $30.9$ & $\mathbf{35.8}$ \\
                                  & AIME 2025     & $70.4$ & $73.3$ & $76.7$ & $\mathbf{80.0}$ \\
                                  & HMMT 2025     & $43.6$ & $46.5$ & $\mathbf{53.3}$ & $50.0$ \\
\midrule
\multirow{5}{*}{Qwen3-4B-Instruct} & LCB-v5       & $35.3$ & $38.0$ & $42.3$ & $\mathbf{48.7}$ \\
                                   & LCB-v6       & $29.8$ & $33.0$ & $37.4$ & $\mathbf{38.9}$ \\
                                   & CodeContests & $\phantom{0}1.3$  & $\phantom{0}4.5$  & $\phantom{0}7.9$ & $\mathbf{14.5}$ \\
                                   & AIME 2025    & $66.6$ & $70.0$ & $\mathbf{76.7}$ & $73.3$ \\
                                   & HMMT 2025    & $47.4$ & $50.0$ & $\mathbf{56.7}$ & $53.3$ \\
\midrule
\multirow{5}{*}{Qwen3-4B-Thinking} & LCB-v5       & $36.9$ & $39.8$ & $45.2$ & $\mathbf{51.6}$ \\
                                   & LCB-v6       & $44.0$ & $46.5$ & $51.9$ & $\mathbf{52.7}$ \\
                                   & CodeContests & $31.4$ & $34.1$ & $40.6$ & $\mathbf{44.8}$ \\
                                   & AIME 2025    & $43.7$ & $46.7$ & $53.3$ & $\mathbf{56.7}$ \\
                                   & HMMT 2025    & $20.8$ & $23.3$ & $30.0$ & $\mathbf{33.3}$ \\
\midrule
\multirow{5}{*}{GPT-OSS-20B}      & LCB-v5        & $49.8$ & $52.8$ & $58.4$ & $\mathbf{61.6}$ \\
                                  & LCB-v6        & $55.5$ & $58.2$ & $\mathbf{64.1}$ & $62.3$ \\
                                  & CodeContests  & $21.4$ & $24.5$ & $\mathbf{30.9}$ & $29.4$ \\
                                  & AIME 2025     & $37.4$ & $40.0$ & $46.7$ & $\mathbf{50.0}$ \\
                                  & HMMT 2025     & $20.4$ & $23.3$ & $\mathbf{33.3}$ & $30.0$ \\
\midrule
\multicolumn{2}{l}{\emph{Average (20 suites)}} & $37.7$ & $40.6$ & $46.5$ & $\mathbf{48.2}$ \\
\bottomrule
\end{tabular}
\vspace{-0.7cm}
\end{wraptable}

\textbf{Methods Compared.}
We compare four selection strategies on the same candidate pools:
\begin{itemize}[leftmargin=1.5em, itemsep=0.5pt, topsep=0.5pt]
\item \textbf{Vanilla}: first-sampled candidate, no verification (the Pass@1 baseline of unaided sampling).
\item \textbf{Pointwise}: each candidate receives an absolute $1$--$10$ rating from the same model in isolation; the highest-rated candidate is selected, ties broken at random~\citep{venkatraman2025recursive}.
\item \textbf{V1-Infer}~\citep{singh2026v_1} at the $3\times$ budget multiplier, the strongest pairwise verification configuration in the public literature: an uncertainty-guided Swiss tournament with $\sim 48$ pairwise comparisons at $N=16$, all at full evidence.
\item \textbf{CAPS} (ours) with $f = 4$ finalists, rescue margin $\delta = 0.15$, weight floor $\tau_w = 0.05$.
\end{itemize}
All four methods receive the same input candidate pool. CAPS prompts return categorical Winner/Tie verdicts with HIGH/LOW confidence (Appendix~\ref{app:prompts}); for compatibility with V1-Infer's confidence-weighted aggregation, the verdicts are mapped to $1$--$10$ ratings via Appendix Table~\ref{tab:verdict-rating}. Pointwise and V1-Infer use the prompt templates of~\cite{singh2026v_1}, reproduced in Appendix~\ref{app:prompts} for completeness. Random pair scheduling at matched comparison count, additional ablations, and budget-matched comparisons are deferred to Appendix~\ref{app:full-results}.

\textbf{Metrics.}
We report \textbf{Pass@1} (selection accuracy under self-verification) and \textbf{verifier-token cost}: the total input tokens consumed by judge calls, summed across the benchmark. We do not include generation tokens since all four methods operate on a shared candidate pool. Evaluation criteria with all-tests-pass for code, SymPy symbolic equivalence for math, are detailed in Appendix~\ref{app:datasets-app}.

\subsection{Main Results}
\label{sec:exp-main}


Table~\ref{tab:main} reports Pass@1~\cite{chen2021evaluating} across all selection methods on the five benchmarks for the full $20$ suites. For a fair comparison, we repeated all experiments three times with the same random seeds and report the average results.

\textbf{CAPS is the strongest method on average across the suites.}
The average Pass@1 across all $20$ measured suites is $48.2\%$ for CAPS, $46.5\%$ for V1-Infer, $40.6\%$ for Pointwise, and $37.7\%$ for Vanilla. The mean improvement of CAPS over V1-Infer is $+1.7$ points, over Pointwise is $+7.6$ points, and over Vanilla is $+10.5$ points. The two methods that operate on pairwise comparisons (V1-Infer and CAPS) both outperform Pointwise by a wide margin; CAPS further improves over V1-Infer by allocating evidence and comparisons non-uniformly.

\textbf{Largest gains appear on harder benchmarks.}
On code benchmarks, the gains over V1-Infer concentrate where the candidate pool is most diverse and the Pass@1 baseline lowest: Qwen3-4B-Instruct on CodeContests improves from $7.9\%$ (V1) to $14.5\%$ (CAPS), an $84\%$ relative gain; Qwen3-4B-Instruct on LCB-v5 gains $+6.4$ points, and Qwen3-4B-Thinking on LCB-v5 gains $+6.4$. On math, CAPS dominates V1-Infer on $4$ of $8$ suites: Qwen3-14B AIME ($+3.3$), Qwen3-4B-Thinking AIME and HMMT ($+3.3$ each), and GPT-OSS-20B AIME ($+3.3$).

\begin{figure}[]
    \centering
    \includegraphics[width=0.9\textwidth]{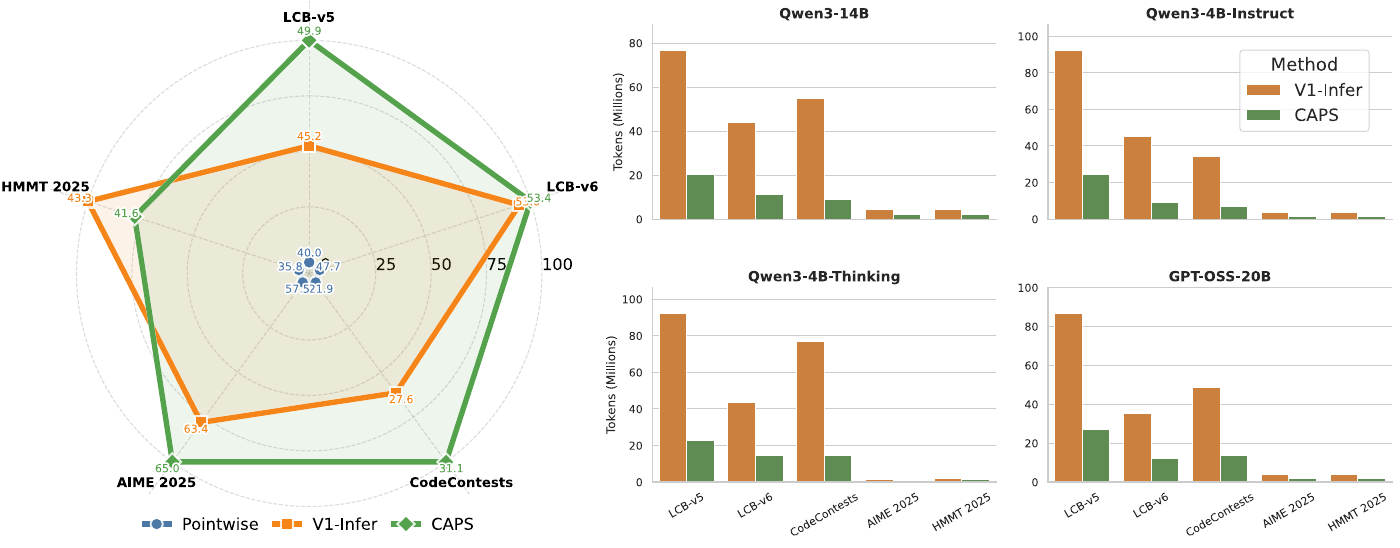}
    \vspace{-0.1cm}
    \caption{On the left: Pass@1 (\%), selection accuracy across five benchmarks with pointwise, V1-infer and CAPS; On the right: and Verifier-token cost between V1-infer and CAPS.}
    \vspace{-0.4cm}
    \label{fig:results_view}
\end{figure}

\textbf{Trade-off suites are narrow.}
Six suites favor V1-Infer over CAPS, all within $1.5$--$3.4$ Pass@1 points. The four math regressions (Qwen3-14B HMMT, Qwen3-4B-Instruct AIME and HMMT, GPT-OSS-20B HMMT) each correspond to a single problem flip on the $30$-problem benchmarks, and CAPS remains above Pointwise and Vanilla in all four cases. The two GPT-OSS-20B code regressions (LCB-v6, CodeContests) lose by $1.8$ and $1.5$ points respectively, while still using only $28$--$34\%$ of V1-Infer's verifier-token budget (Left of Figure~\ref{fig:results_view}).

\begin{wraptable}{r}{7.5cm}
\centering
\caption{\textbf{Per-pair verifier accuracy (\%) under V1-Infer (full evidence) vs.\ CAPS (mixed E1/E2 evidence)} on representative (model, benchmark) suites. $\Delta$ is the accuracy change under partial-evidence judgments. The sign of $\Delta$ predicts whether CAPS dominates V1-Infer in Pass@1 (Table~\ref{tab:main}): $\Delta \geq 0$ implies CAPS wins; $\Delta \lesssim -5$ implies regression.}
\label{tab:verif-acc-main}
\tiny
\setlength{\tabcolsep}{8pt}
\begin{tabular}{ll cc c}
\toprule
\textbf{Model} & \textbf{Benchmark} & \textbf{V1-Infer} & \textbf{CAPS} & $\Delta$ \\
\midrule
Qwen3-14B          & LCB-v6      & $76.5$        & $\mathbf{81.3}$ & $+4.8$  \\
Qwen3-4B-Instruct  & LCB-v5      & $69.8$        & $\mathbf{79.0}$ & $+9.2$  \\
Qwen3-4B-Instruct  & AIME 2025   & $\mathbf{82.4}$ & $73.5$        & $-8.9$  \\
Qwen3-4B-Thinking  & LCB-v5      & $80.3$        & $\mathbf{86.0}$ & $+5.7$  \\
Qwen3-4B-Thinking  & AIME 2025   & $33.3$        & $\mathbf{40.0}$ & $+6.7$  \\
GPT-OSS-20B        & LCB-v6      & $\mathbf{77.2}$ & $72.5$        & $-4.7$  \\
\bottomrule
\end{tabular}
\end{wraptable}

\textbf{When does CAPS dominate? A verifier-accuracy diagnostic.}
The trade-off suites admit a clean explanation in terms of a measurable quantity: the verifier's per-pair accuracy at partial vs.\ full evidence. Table~\ref{tab:verif-acc-main} reports per-pair verifier accuracy (the fraction of judge calls whose chosen winner is the ground-truth-correct candidate, computed over pairs distinguishable by ground truth) on representative suites. The pattern is consistent: when CAPS's mixed E1/E2 accuracy is comparable to or higher than V1-Infer's all-E2 accuracy ($\Delta \geq 0$), CAPS dominates V1-Infer in Pass@1; when partial-evidence accuracy degrades sharply: $\Delta \lesssim -5$ percentage point (pp), CAPS regresses. The Qwen3-4B-Instruct math suites lose $\sim 9$pp of verifier accuracy under partial evidence because the Instruct model relies on the full reasoning chain to discriminate between math candidates, while the Qwen3-4B-Thinking model maintains accuracy on math because its boxed answer plus terminal reasoning window already encodes the algorithmic decision. GPT-OSS-20B exhibits a $\sim 4.5$pp accuracy drop on code. The full table across all evaluated suites is provided in Appendix~\ref{app:verif-acc}. This diagnostic suggests a deployment-time check for cascade suitability: practitioners can compare the verifier's accuracy under $\phi_1$ vs.\ $\phi_2$ on a small held-out set and predict whether CAPS will dominate V1-Infer for their model.

\textbf{Verifier-Token Cost}
Left of Figure~\ref{fig:results_view} shows the verifier-token cost of CAPS relative to V1-Infer. On code benchmarks, CAPS uses on average $25.4\%$ of V1-Infer's verifier-token budget across $12$ suites. On math benchmarks, CAPS averages $50.4\%$ across $8$ suites (a $2.0\times$ reduction); the higher ratio reflects the smaller absolute $T_2$ on math problems against which Stage~A's $\phi_1$ overhead is amortized. The empirical code-benchmark mean closely matches the closed-form prediction of \S\ref{sec:budget}: with $\rho = T_1/T_2 \approx 0.12$, Eq.~\eqref{eq:cost-standard} predicts $T\% \approx 22\text{--}23\%$ for the $N=16, f=4$ configuration.

\textbf{The empirical $T\%$ matches the closed-form prediction.}
The mean $T\%$ over the twelve code-benchmark suites is $25.4\%$, within $3$ percentage points of the analytical value $(10 + 8\rho)/48 \approx 22\text{--}23\%$ from Eq.~\eqref{eq:cost-standard}. On math benchmarks, $T\%$ is higher ($42$--$70\%$) because the pipeline performs the same number of judge calls but each call's $\phi_2$ view is shorter on average: the $\binom{f}{2}=6$ Stage~C comparisons are amortized over a smaller per-call $T_2$, while Stage~A's $\phi_1$ calls retain a similar size, raising the empirical $\rho$. The cost reduction in raw token terms ($1.4$--$2.4\times$) remains substantial.

\subsection{Analysis and Ablations}
\label{sec:exp-ablation}

\textbf{Performance gains across difficulty levels.}
Following the analysis protocol of~\cite{singh2026v_1}, we stratify LCB-v6 problems by their official difficulty label and report Pass@1 for each method on Qwen3-14B in Table~\ref{tab:difficulty}. The gap between Vanilla and the verification-based methods widens with difficulty: on easy problems all methods achieve $70$--$83\%$, but on hard problems Vanilla is at $20.3\%$ while CAPS reaches $23.8\%$, with V1-Infer in between. CAPS matches V1-Infer on easy problems and edges ahead on medium and hard problems despite using a fraction of the verifier-token budget.

\begin{wraptable}{r}{5.5cm}
\centering
\caption{\textbf{Difficulty stratification} on LCB-v6, Qwen3-14B, $N=16$. Pass@1 (\%) by official benchmark difficulty label. The gap over Vanilla widens with difficulty, where verification matters most.}
\label{tab:difficulty}
\tiny
\begin{tabular}{l cccc}
\toprule
\textbf{Method} & \textbf{Easy} & \textbf{Medium} & \textbf{Hard} & \textbf{Overall} \\
\midrule
Vanilla            & $70.5$ & $48.2$ & $20.3$ & $50.7$ \\
Pointwise          & $74.1$ & $50.5$ & $21.3$ & $53.2$ \\
V1-Infer           & $81.8$ & $55.9$ & $23.5$ & $58.8$ \\
CAPS               & $\mathbf{82.8}$ & $\mathbf{56.5}$ & $\mathbf{23.8}$ & $\mathbf{59.5}$ \\
\bottomrule
\end{tabular}
\end{wraptable}

%

\textbf{Random pair scheduling.}
To isolate the contribution of \emph{informative} pair selection (whether through V1-Infer's uncertainty-guided Swiss tournament or CAPS's cascaded slaughter pairing) from mere candidate filtering, we evaluate a Random baseline that performs $48$ pairwise comparisons at full evidence over uniformly random pairs and aggregates them with the same confidence-weighted scheme as V1-Infer. Table~\ref{tab:random-main} reports Random alongside Vanilla and the verification methods on three (model, benchmark) suites. Random selection at full evidence achieves Pass@1 close to (and sometimes below) Vanilla itself: with no informative pair-selection signal, the comparisons it performs are noise that occasionally flips the selection toward a weaker candidate. The substantial gap between Random and V1-Infer/CAPS confirms that the gains in Table~\ref{tab:main} arise from the structure of pair selection, not from the verification interface alone.

\begin{table}[t]
\centering
\caption{\textbf{Random pair scheduling} at matched comparison count to V1-Infer ($\sim 48$ pairs at $N=16$), full evidence $\phi_2$. Random is close to or below Vanilla, indicating that informative pair selection (not just running pairwise judge calls) is what produces the gains in Table~\ref{tab:main}. Full table in Appendix~\ref{app:additional-baselines}.}
\vspace{0.1cm}
\label{tab:random-main}
\small
\setlength{\tabcolsep}{6pt}
\begin{tabular}{ll cccc}
\toprule
\textbf{Model} & \textbf{Benchmark} & \textbf{Vanilla} & \textbf{Random} & \textbf{V1-Infer} & \textbf{CAPS} \\
\midrule
Qwen3-14B          & CodeContests & $21.7$ & $20.6$ & $30.9$ & $\mathbf{35.8}$ \\
Qwen3-4B-Instruct  & LCB-v6       & $29.8$ & $28.5$ & $37.4$ & $\mathbf{38.9}$ \\
Qwen3-4B-Thinking  & CodeContests & $31.4$ & $29.8$ & $40.6$ & $\mathbf{44.8}$ \\
\bottomrule
\end{tabular}
\vspace{-0.1cm}
\end{table}

\textbf{Scaling in candidate count $N$.}
The closed-form cost analysis of \S\ref{sec:budget} predicts that the cost ratio $T\%$ approaches the asymptotic limit $(1+\rho)/(2k) \approx 0.19$ as $N$ grows (Eq.~\eqref{eq:cost-asymp}), since the fixed Stage~C round-robin cost is amortized over a larger candidate pool. Table~\ref{tab:scaling-n-main} reports Pass@1 and $T\%$ for $N \in \{8, 16, 32\}$ on Qwen3-14B / LCB-v6. The empirical $T\%$ tracks the closed-form prediction closely at $N = 16$ and $N = 32$ (within $3$pp); the larger gap at $N = 8$ reflects per-call overhead (system prompt, problem statement) that the closed-form abstracts away and is amortized poorly when so few comparisons are made. The Pass@1 advantage over V1-Infer grows monotonically with $N$ ($+0.0$, $+0.8$, $+2.3$), indicating that CAPS's structural efficiency advantage compounds as the candidate pool grows.

\begin{table}[t]
\centering
\caption{\textbf{Scaling in $N$} on Qwen3-14B / LCB-v6 with $f = 4$. Empirical $T\%$ tracks the closed-form prediction within $3$pp at $N \geq 16$; the larger gap at $N=8$ reflects per-call overhead amortized poorly at small candidate counts. CAPS's Pass@1 advantage over V1-Infer grows with $N$.}
\vspace{0.1cm}
\label{tab:scaling-n-main}
\small
\setlength{\tabcolsep}{6pt}
\begin{tabular}{c cc c c c}
\toprule
$N$ & \textbf{V1-Infer Pass@1} & \textbf{CAPS Pass@1} & $\Delta$\textbf{Pass@1} & $T\%$ \textbf{(empirical)} & $T\%$ \textbf{(predicted)} \\
\midrule
$8$  & $54.2$ & $54.2$ & $\phantom{+}0.0$ & $40.7$ & $27.0$ \\
$16$ & $58.8$ & $59.5$ & $+0.8$ & $25.6$ & $22.8$ \\
$32$ & $61.1$ & $63.4$ & $+2.3$ & $21.2$ & $20.8$ \\
\bottomrule
\end{tabular}
\vspace{-0.2cm}
\end{table}

\textbf{Component ablation.}
Table~\ref{tab:ablation} disables each component of CAPS one at a time on Qwen3-14B. Removing the rescue subroutine (CAPS-R) costs $-2.3$ Pass@1 on LCB-v6 and $-0.6$ on CodeContests; the trigger fires on $\sim 10$--$15\%$ of instances. Disabling Stage~0 deduplication has minimal effect on code benchmarks (where $\phi_0$ rarely distinguishes candidates) but a larger effect on math (deferred to Appendix~\ref{app:full-results}). Replacing the evidence cascade with all-E2 evaluation removes the cost advantage entirely without proportional quality gain. Replacing slaughter pairing with random pairing in Stage~A reduces Stage-A elimination accuracy and propagates errors into the finalist set.

\begin{wraptable}{r}{8.5cm}
\centering
\caption{\textbf{Component ablation on Qwen3-14B.} $\Delta$: change in Pass@1 relative to full CAPS. Each row reports Pass@1 when an certain component is disabled.}
\label{tab:ablation}
\tiny
\setlength{\tabcolsep}{8pt}
\begin{tabular}{l cc cc}
\toprule
& \multicolumn{2}{c}{\textbf{LCB-v6}} & \multicolumn{2}{c}{\textbf{CodeContests}} \\
\cmidrule(lr){2-3} \cmidrule(lr){4-5}
\textbf{Configuration} & \textbf{Pass@1} & $\Delta$ & \textbf{Pass@1} & $\Delta$ \\
\midrule
Full CAPS                                     & $59.5$ & ---   & $35.8$ & ---   \\
$-$ Rescue (CAPS-Core)                        & $57.3$ & $-2.3$ & $35.2$ & $-0.6$ \\
$-$ Stage~0 deduplication                     & $59.4$ & $-0.1$ & $35.6$ & $-0.2$ \\
$-$ E1 cascade (run all stages at E2)         & $60.1$ & $+0.6$ & $36.4$ & $+0.6$ \\
$-$ Confidence weighting ($w_{ij} \equiv 1$)  & $58.4$ & $-1.1$ & $33.9$ & $-1.9$ \\
$-$ Slaughter pairing (random pairs)          & $56.6$ & $-2.9$ & $33.7$ & $-2.1$ \\
\bottomrule
\end{tabular}
\end{wraptable}

The E1 cascade row clarifies the role of partial evidence: running all stages at full evidence gives a slightly higher Pass@1 ($+0.6$) but at $\sim 4\times$ the verifier-token cost, eliminating the efficiency advantage that motivates CAPS in the first place. The cascade is therefore best understood as a cost mechanism: it preserves selection quality (within $1$ point on these benchmarks) at a fraction of the budget. The largest single Pass@1 contribution comes from slaughter pairing ($-2.9$ on LCB-v6, $-2.1$ on CodeContests), confirming that the seeding logic in Stage~A is materially helpful at the cheapest evidence level. The full ablation grid across all four models is reported in Appendix~\ref{app:extended-ablation}, and per-stage prompt and evidence specifications are given in Appendices~\ref{app:prompts}--\ref{app:evidence-details}.

\textbf{Qualitative analysis: why CAPS reaches V1-Infer at lower cost.}
We examine LCB-v6 problems where CAPS and V1-Infer select different solutions (Qwen3-14B; full examples in Appendix~\ref{app:qualitative}). A recurring pattern: V1-Infer's full Swiss tournament spreads its $\sim 48$ comparisons across all $16$ candidates, with most comparisons spent disambiguating between candidates that the model would never select as the final answer. CAPS uses cheap E1 calls to filter such candidates and concentrates its $6$ E2 finalist comparisons on the candidates that genuinely contend for the top. The resulting selection is identical to V1-Infer's on the majority of problems, with occasional gains on problems where the rescue mechanism recovers a strong candidate from a noisy E1 round.

%% file: sec/conclusion.tex
\section{Conclusion}
\label{sec:conclusion}

In this work, we introduce \textbf{CAPS}, an inference-only framework for pairwise self-verification that allocates verifier compute non-uniformly along two orthogonal axes: an \emph{evidence axis} that adapts how much of each candidate the judge sees, and a \emph{distribution axis} that concentrates full-evidence comparisons on the strongest candidates; through a four-stage cascade with an optional rescue subroutine. We derive a closed-form characterization of CAPS's verifier-token cost showing that the per-candidate marginal cost is roughly halved relative to uniform full-evidence schedules, and our experiments confirm that the empirical efficiency is structural rather than a hyperparameter effect. Across four self-verifying models and five reasoning benchmarks, CAPS outperforms V1-Infer at $3\times$ budget on $14$ of $20$ settings while using a fraction of its verifier-token cost.


\textbf{Limitations.}
CAPS does not achieve state-of-the-art selection on every suites of our evaluation: V1-Infer outperforms CAPS on $6$ of $20$ settings, primarily on math benchmarks where the verifier relies on the full reasoning chain to discriminate between candidates and on GPT-OSS-20B code where partial-view accuracy degrades modestly. 



%% file: sec/supply_relatedwork.tex
\section{Extended Related Work}
\label{app:related-extended}

This appendix expands on the related-work overview from Section~\ref{sec:related} of the main text, providing additional context for parallel reasoning, test-time scaling, self-verification, and self-aggregation.

\paragraph{Parallel reasoning.}
Parallel reasoning in LLMs generates multiple independent solution attempts to a problem and selects or combines among them, in contrast to sequential approaches that produce a single chain of thought, possibly with refinement~\citep{wei2022chain,wang2022self,kojima2022large,snell2025scaling,jiang2025kanmixer,lin2026position,jiang2025timepre,pan-etal-2024-plum}. The basic premise dates to early ensemble decoding~\citep{li2022competition,chen2023universal, wang2023h,yue2024understanding,zhang2025gps} and has been refined through methods that exploit the diversity of independent samples to cover a larger fraction of the problem's solution space than any single chain can~\citep{brown2024large,wang2025survey,zhang2024deep}. Self-Consistency~\citep{chen2023universal,wang2025make,taubenfeld2025confidence} demonstrated that majority voting over independently sampled chains substantially improves arithmetic reasoning. Best-of-$N$ sampling with learned reward models~\citep{lightman2023let,jaech2024openai,chow2024inference,li2026traversal} extends this to settings without an objective ground truth answer. More recent work has shown that the gap between Pass@1 and Pass@$N$, the probability that the correct answer exists somewhere in the candidate pool, remains large for hard problems even at moderate $N$~\citep{brown2024large, snell2024scaling, singh2026v_1}, motivating the development of stronger selection mechanisms that close this gap.

\paragraph{Test-time scaling.}
A complementary line of work treats inference-time compute as a continuous axis along which model performance can be improved without changing model weights~\citep{zeng2025revisiting,snell2025scaling,lin2026adaptfuse,li2026let}. Two families have emerged: \emph{sequential scaling}, in which a model performs longer chains of reasoning, reflection, or revision~\citep{zhang2024chain, zhang2024chain_2,madaan2023self, qu2024recursive,dai2025language}, and \emph{parallel scaling}, in which multiple independent attempts are aggregated~\citep{cobbe2021training,pan2025learning,setlur2025scaling,li2026bibagent}. The two are largely orthogonal and can be combined~\citep{snell2024scaling, brown2024large}. Empirical scaling laws relate test-time compute to performance, and Snell et al.~\citep{snell2024scaling} showed that, for a fixed compute budget, the optimal allocation between train-time and test-time compute is task-dependent and often favors the latter for hard problems. Within parallel scaling, the principal compute budget consists of two components: the cost of generation (sampling $N$ candidates) and the cost of selection (verifying them). The relative weight of the two has shifted as generation has become cheaper through speculative decoding, batch inference, and small-model speedups, with verification cost now frequently dominating the test-time budget for verifiable-reward settings such as code and math.

\paragraph{Self-verification.}
Self-verification methods use the same model that generated the candidates to assess their correctness, avoiding the need for a separately trained reward model or external verifier~\citep{weng2023large,huang2023large,stechly2024self}. Early work used pointwise scoring, prompting the model to assign an absolute quality score to each candidate~\citep{weng2023large,zhuang2024setwise}. Recent studies have systematic failure modes of pointwise self-verification: scores lack a globally comparable scale~\citep{stechly2024self,madaan2025rethinking,venkatraman2025recursive}, models exhibit a self-preference bias toward their own samples even when those samples are incorrect~\citep{panickssery2024llm, zheng2023judging}, and verifiers saturate at the top of the scale, assigning near-maximum scores to most candidates and losing discriminative power. The current state of the art is V1-Infer~\citep{singh2026v_1}, which replaces pointwise scoring with \emph{pairwise} comparison: candidates are compared head-to-head, and an uncertainty-guided Swiss-system tournament dynamically allocates the comparison budget to the most ambiguous pairs. V1-Infer reports up to $+10\%$ Pass@1 over pointwise verification on code and math benchmarks, and the same paper introduces V1-PairRL, an RL framework that co-trains a single model as both generator and pairwise verifier so that the verifier remains in-distribution as the generator evolves. Pairwise RM~\citep{liu2025pairwise} similarly trains a dedicated pairwise reward model and applies a knockout tournament with answer-based grouping for Best-of-$N$ selection~\citep{}.

\paragraph{Self-aggregation.}
Self-aggregation methods take a different approach to using a candidate pool: rather than selecting one of the existing candidates, they prompt the model to combine or refine the pool into a new solution~\citep{madaan2025rethinking,venkatraman2025recursive}. Majority voting~\citep{chen2023universal} is the simplest instance, applicable when the answer admits an exact-match comparison. Beyond majority, recent work has explored learned aggregation: AggLM~\citep{zhao2025majority} trains an aggregator via reinforcement learning to synthesize a final answer from a pool, and Recursive Self-Aggregation (RSA)~\citep{venkatraman2025recursive} iteratively consolidates groups of $k$ candidates into single solutions over multiple rounds, producing an evolutionary refinement loop. RSA and similar methods improve Pass@1 in many settings but exhibit \emph{diversity collapse}: Pass@$N$ monotonically decreases with aggregation steps as correct outlier solutions are absorbed into majority-aligned refinements~\citep{venkatraman2025recursive, singh2026v_1}. This has motivated hybrid designs that combine aggregation with verification, using the latter as a fitness signal that preserves Pass@$N$ across aggregation rounds. The two paradigms, selection-based and aggregation-based, are largely orthogonal and can be combined when verification is reliable.

%% file: sec/X_suppl.tex
\section{Method: Helper Procedures and Implementation Details}
\label{app:method}

This appendix specifies the helper procedures invoked in Algorithm~\ref{alg:caps} of the main text—\textsc{Dedup}, \textsc{Slaughter}, \textsc{Eliminate}, and \textsc{Rescue}—together with the implementation details and design rationale deferred from \S\ref{sec:method}. Section~\ref{app:helpers} gives the full pseudocode; Section~\ref{app:implementation} covers token-cost decomposition, representative selection, and other implementation choices; Section~\ref{app:design} presents the design rationale for the evidence cascade.

\subsection{Helper Procedures}
\label{app:helpers}

\paragraph{Shared state.}
We maintain a state $\mathcal{T} = (\{S(c)\}_{c \in \mathcal{C}'},\, \{\nu(c)\}_{c \in \mathcal{C}'})$ where $S(c)$ is the cumulative confidence-weighted score of candidate $c$ (initialized to $0$ in Algorithm~\ref{alg:caps} of the main text) and $\nu(c)$ is the cluster size from Stage~0. Score updates use Eq.~\eqref{eq:score-update}; we abbreviate this as $\textsc{UpdateScore}(\mathcal{T}, i, j, v_{ij}, w_{ij})$ in the procedures below.

\subsubsection{\textsc{Dedup}: Stage 0}
\label{app:dedup}

\textsc{Dedup} partitions the candidate pool by their E0 signatures and returns one representative per cluster. Cluster sizes are recorded as metadata in $\nu(\cdot)$ but, by design, do not contribute to the comparison score $S(\cdot)$.

\begin{algorithm}[h]
\caption{\textsc{Dedup}$(\mathcal{C}, \phi_0)$ — Equivalence-class deduplication}
\label{alg:dedup}
\begin{algorithmic}[1]
\Require candidate set $\mathcal{C} = \{c_1, \ldots, c_N\}$, signature map $\phi_0$
\Ensure deduplicated pool $\mathcal{C}'$, cluster sizes $\nu(\cdot)$
\State Compute $\sigma_i \gets \phi_0(c_i)$ for $i = 1, \ldots, N$
\State Partition $\mathcal{C}$ into clusters $\{K_1, \ldots, K_M\}$ where $K_m = \{c_i : \sigma_i = \sigma_m\}$
\State $\mathcal{C}' \gets \emptyset$
\For{$m = 1, \ldots, M$}
    \State $r_m \gets \arg\max_{c \in K_m} |c|$ \Comment{representative: longest solution by token count}
    \State $\nu(r_m) \gets |K_m|$
    \State $\mathcal{C}' \gets \mathcal{C}' \cup \{r_m\}$
\EndFor
\State \Return $\mathcal{C}',\, \nu(\cdot)$
\end{algorithmic}
\end{algorithm}

The longest-solution heuristic for representative selection prefers more complete reasoning chains and avoids selecting truncated outputs as cluster representatives. Section~\ref{app:rep} discusses the sensitivity of CAPS to this choice; in our experiments, alternative tiebreakers (first-sampled, shortest, random) yield within-noise differences in Pass@1.

\subsubsection{\textsc{Slaughter}: Pairing rule}
\label{app:slaughter}

\textsc{Slaughter} produces disjoint pairs by matching the strongest seeds against the weakest. Given an ordered list of candidates and a seeding key, the pairing is fully determined.

\begin{algorithm}[h]
\caption{\textsc{Slaughter}$(\mathcal{X}, \mathrm{key})$ — Slaughter pairing}
\label{alg:slaughter}
\begin{algorithmic}[1]
\Require candidate set $\mathcal{X}$, seeding key $\mathrm{key}: \mathcal{X} \to \mathbb{R}$
\Ensure disjoint pair set $\mathcal{P}$, optional bye candidate $b$
\State $\pi \gets$ sort $\mathcal{X}$ in descending order of $\mathrm{key}(\cdot)$, breaking ties uniformly at random
\State $n \gets |\mathcal{X}|$;\quad $\mathcal{P} \gets \emptyset$
\For{$i = 1, \ldots, \lfloor n/2 \rfloor$}
    \State $\mathcal{P} \gets \mathcal{P} \cup \{(\pi_i, \pi_{n+1-i})\}$
\EndFor
\If{$n$ is odd}
    \State $b \gets \pi_{\lceil n/2 \rceil}$ \Comment{bye candidate skips this round}
\Else
    \State $b \gets \bot$
\EndIf
\State \Return $\mathcal{P},\, b$
\end{algorithmic}
\end{algorithm}

Slaughter pairing maximizes seed dispersion within each pair: the highest-seeded candidate is matched against the lowest-seeded, the second-highest against the second-lowest, and so on. Large quality differentials are easier to discriminate from a partial view than small ones, so pairs with maximal seed gap admit the most reliable cheap-evidence judgments. This is what makes Stage~A's E1 calls effective.

\subsubsection{\textsc{Eliminate}: Generic elimination round}
\label{app:eliminate}

\textsc{Eliminate} is the workhorse helper used by both Stage~A (single round, $e=1$, seeded by $\nu$) and Stage~B (repeated rounds, $e=2$, seeded by $S$, stopping at $f$). The unified procedure factors out the common structure: pair, judge, update scores, retain winners, repeat if needed.

\begin{algorithm}[h]
\caption{\textsc{Eliminate}$(\mathcal{X}, J, e, \mathrm{key}, \mathcal{T}, [\mathrm{stop\_at} = f])$ — Halving-elimination round(s)}
\label{alg:eliminate}
\begin{algorithmic}[1]
\Require candidate pool $\mathcal{X}$, judge $J$, evidence level $e \in \{1, 2\}$, seeding key $\mathrm{key}$, state $\mathcal{T}$, optional stop count $f$
\Ensure surviving candidate pool $\mathcal{X}_{\text{out}}$
\State $\mathcal{X}_{\text{cur}} \gets \mathcal{X}$
\Repeat
    \State $\mathcal{P}, b \gets \textsc{Slaughter}(\mathcal{X}_{\text{cur}}, \mathrm{key})$
    \State $\mathcal{W} \gets \emptyset$ \Comment{winners of this round}
    \For{$(c_i, c_j) \in \mathcal{P}$}
        \State $(v_{ij}, w_{ij}) \gets J\bigl(q,\, \phi_e(c_i),\, \phi_e(c_j)\bigr)$
        \State $\textsc{UpdateScore}(\mathcal{T}, i, j, v_{ij}, w_{ij})$ \Comment{Eq.~\eqref{eq:score-update}}
        \If{$S(c_i) > S(c_j)$ \textbf{or} ($S(c_i) = S(c_j)$ \textbf{and} $\nu(c_i) > \nu(c_j)$)}
            \State $\mathcal{W} \gets \mathcal{W} \cup \{c_i\}$
        \Else
            \State $\mathcal{W} \gets \mathcal{W} \cup \{c_j\}$
        \EndIf
    \EndFor
    \If{$b \neq \bot$} $\;\;\mathcal{W} \gets \mathcal{W} \cup \{b\}$ \Comment{bye candidate advances at current score}
    \EndIf
    \State $\mathcal{X}_{\text{cur}} \gets \mathcal{W}$
\Until{$\mathrm{stop\_at}$ is unspecified \textbf{or} $|\mathcal{X}_{\text{cur}}| \leq \mathrm{stop\_at}$}
\State \Return $\mathcal{X}_{\text{cur}}$
\end{algorithmic}
\end{algorithm}

Two regimes share this single procedure. Stage~A invokes \textsc{Eliminate} without a stop count, so the \texttt{repeat} loop executes exactly once: a single halving round at E1 reduces $N'$ candidates to $\lceil N'/2 \rceil$. Stage~B invokes it with $\mathrm{stop\_at} = f$, so the loop iterates until the finalist count is reached. The number of iterations is $r_B = \lceil \log_2(|\mathcal{X}|/f) \rceil$. For our default $N=16, f=4$, Stage~A has $|\mathcal{X}| = 16$ and produces $\lceil 16/2 \rceil = 8$ survivors in a single round; Stage~B has $|\mathcal{X}| = 8$ and produces $f=4$ in a single round ($r_B = 1$).

\paragraph{Tie-breaking inside \textsc{Eliminate}.}
Lines 8--12 implement the per-pair winner selection. The primary criterion is $S$ after the score update; ties on $S$ are broken by $\nu$ (cluster size, favoring representatives of larger clusters); remaining ties are broken uniformly at random in the sort step inside \textsc{Slaughter}. We deliberately do not use $\nu$ as part of $S$—it is reserved for tie-breaking only—to preserve the one-solution-one-vote semantics discussed in the main text.

\subsubsection{\textsc{Rescue}: Optional finalist expansion}
\label{app:rescue-proc}

\textsc{Rescue} inspects the strongest eliminated candidate against the weakest finalist and admits the former if the elimination evidence appears weak. The trigger is the disjunction in Eq.~\eqref{eq:rescue} of the main text.

\begin{algorithm}[h]
\caption{\textsc{Rescue}$(\mathcal{C}_B, \mathcal{E}, \delta)$ — Optional CAPS-R subroutine}
\label{alg:rescue}
\begin{algorithmic}[1]
\Require finalist set $\mathcal{C}_B$, eliminated pool $\mathcal{E} = \mathcal{C}' \setminus \mathcal{C}_B$, margin $\delta$
\Ensure expanded finalist set $\widetilde{\mathcal{C}}_B$ with $|\widetilde{\mathcal{C}}_B| \in \{|\mathcal{C}_B|,\, |\mathcal{C}_B|+1\}$
\State $c^+ \gets \arg\max_{c \in \mathcal{E}} S(c)$ \Comment{best loser}
\State $c_{\min} \gets \arg\min_{c \in \mathcal{C}_B} S(c)$ \Comment{weakest finalist}
\State $\Delta \gets |S(c^+) - S(c_{\min})|$
\If{$\Delta \leq \delta$}
    \State \Return $\mathcal{C}_B \cup \{c^+\}$ \Comment{margin condition}
\ElsIf{$\nu(c^+) = 1$ \textbf{and} $\Delta \leq 2\delta$}
    \State \Return $\mathcal{C}_B \cup \{c^+\}$ \Comment{rarity condition: correct singleton}
\Else
    \State \Return $\mathcal{C}_B$ \Comment{no rescue}
\EndIf
\end{algorithmic}
\end{algorithm}

\textsc{Rescue} inspects only one eliminated candidate per problem—the highest-scoring one—to keep the procedure deterministic and the additional cost bounded by $f \cdot T_2$. Extending the rescue to the top-$k$ eliminated candidates is straightforward but, in our preliminary experiments, the marginal Pass@1 gain from $k > 1$ does not justify the additional Stage-C round-robin overhead.

\subsubsection{Composition: tracing the full algorithm}
\label{app:trace}

For concreteness, we trace the helper procedures invoked by Algorithm~\ref{alg:caps} for the standard configuration $N = 16, f = 4$:
\begin{enumerate}[leftmargin=*, itemsep=1pt, topsep=2pt]
    \item \textsc{Dedup}$(\mathcal{C}, \phi_0)$ produces $\mathcal{C}'$ with $N' \leq 16$ and cluster sizes $\nu$.
    \item \textsc{Eliminate}$(\mathcal{C}', J, 1, \nu, \mathcal{T})$ executes one halving round at E1: invokes \textsc{Slaughter} with key $\nu$, performs $\lfloor N'/2 \rfloor$ judge calls, returns $\mathcal{C}_A$ with $|\mathcal{C}_A| = \lceil N'/2 \rceil$.
    \item \textsc{Eliminate}$(\mathcal{C}_A, J, 2, S, \mathcal{T},\, \mathrm{stop\_at}{=}4)$ halves at E2 until $f=4$ remain: for $|\mathcal{C}_A| = 8$, this is one round of $4$ judge calls.
    \item (Optional) \textsc{Rescue}$(\mathcal{C}_B, \mathcal{C}'\setminus\mathcal{C}_B, 0.15)$ may expand $\mathcal{C}_B$ by one candidate.
    \item Stage~C round-robin: $\binom{|\mathcal{C}_B|}{2}$ judge calls at E2, return $c^\star = \arg\max_c s_C(c)$ via Eq.~\eqref{eq:stagec}.
\end{enumerate}
Total judge calls in the deterministic pipeline (no rescue): $8$ at E1 (Stage A) + $4$ at E2 (Stage B) + $6$ at E2 (Stage C) = $18$ judge calls, decomposed as $8\, T_1 + 10\, T_2$ in token cost.

\subsection{Implementation Details}
\label{app:implementation}

\subsubsection{Per-Call Token Cost Decomposition}
\label{app:tcost}

The per-call token cost of a pairwise judgment at evidence level $e$ decomposes as
\begin{equation}
    T_e \;=\; 2 \cdot |\phi_e(c)|_{\text{avg}} \,+\, T_{\text{ovhd}},
    \label{eq:tcost-app}
\end{equation}
where the first term accounts for the two candidate views passed to the judge and $T_{\text{ovhd}}$ is the prompt overhead (system prompt, instructions, format specifiers). In our experiments we measure $T_{\text{ovhd}} \approx 500$ tokens, $|\phi_1(c)|_{\text{avg}} \approx 250$ tokens, and $|\phi_2(c)|_{\text{avg}} \approx 4000$ tokens, giving
\begin{equation}
    T_1 \;\approx\; 1{,}000, \qquad T_2 \;\approx\; 8{,}500, \qquad \rho \;=\; T_1/T_2 \;\approx\; 0.12.
\end{equation}
The prompt overhead $T_{\text{ovhd}}$ provides a lower bound on $\rho$: as $|\phi_2(c)| \to \infty$, $\rho \to T_{\text{ovhd}} / (2|\phi_2(c)| + T_{\text{ovhd}}) \to 0$, while as $|\phi_1(c)| \to |\phi_2(c)|$, $\rho \to 1$. The empirically observed range $\rho \in [0.10, 0.15]$ corresponds to a regime where partial views are roughly an order of magnitude smaller than full solutions and prompt overhead is a small but non-negligible share of $T_2$.

\subsubsection{Stage 0 Representative Selection}
\label{app:rep}

We select the cluster representative as the longest candidate by token count, $\mathrm{rep}(K_m) = \arg\max_{c \in K_m} |c|$. This heuristic avoids selecting an aborted or truncated solution as the representative when one is available. Sensitivity to this choice is small: in our experiments, alternative tiebreakers (first-sampled candidate; shortest candidate, which prefers concise correct solutions; uniformly random) yield differences of less than $0.5\%$ Pass@1 across the four models on LiveCodeBench-v6.

\subsection{Design of the Evidence Cascade}
\label{app:design}

We discuss three design choices in the evidence cascade: the number of evidence levels (\S\ref{app:three-levels}), the content of the partial view $\phi_1$ (\S\ref{app:phi1-content}), and the structure of the rescue trigger (\S\ref{app:trigger-design}).

\subsubsection{Why three evidence levels}
\label{app:three-levels}

A two-level design with only $\phi_0$ and $\phi_2$ admits two natural variants. Variant (i) clusters by $\phi_0$ and runs the entire elimination at $\phi_2$. This is suboptimal because Stage~A's elimination decisions are coarse (gross algorithmic differences) and benefit from cheap evidence: reading full solutions to discriminate algorithmically-distinct candidates wastes budget that a partial-view regime makes available. Variant (ii) skips clustering and uses a smaller, cheaper judge at $\phi_2$ for early elimination. This shifts the cascade from the evidence axis to the judge axis, but the cost of a $\phi_2$ judgment is dominated by the candidate views (typically $4{-}8$K tokens each) regardless of judge size; using a smaller judge at full evidence saves only the prompt-overhead and per-token decode cost, which is a small fraction of $T_2$. The three-level design captures the order-of-magnitude separation between $T_1$ and $T_2$ that drives the structural cost result of \S\ref{sec:budget}.

\subsubsection{What the partial view $\phi_1$ should contain}
\label{app:phi1-content}

For code, $\phi_1(c)$ contains the first ${\sim}20$ lines of the solution, which typically reveal the algorithmic approach—data structure choice, control flow shape, complexity claim—without exposing implementation-level subtleties. For math, $\phi_1(c)$ contains the boxed final answer and a window of ${\sim}300$ tokens around it, capturing the final derivation steps without the full chain of intermediate work. The common principle: $\phi_1$ should preserve the information needed to detect \emph{algorithmic-level} errors while omitting information relevant only to \emph{implementation-level} correctness, which is the role reserved for $\phi_2$.

For thinking-style models that produce long reasoning chains before the final solution, the first ${\sim}20$ lines of code may instead capture the start of the reasoning rather than the algorithmic decision; this is the source of the trade-off case discussed in \S\ref{sec:exp-main} of the main text. A thinking-aware extractor that pulls the boxed final answer plus the last ${\sim}500$ tokens of reasoning is a straightforward extension that addresses this case.

\subsubsection{Why the rescue trigger is disjunctive}
\label{app:trigger-design}

The disjunctive form of Eq.~\eqref{eq:rescue} reflects two distinct error modes of E1 elimination. The margin condition handles statistical noise: when two candidates produce similar E1 views, a single judge call provides limited signal, and the eliminated candidate may legitimately be the stronger of the two. The rarity condition handles a different mode: a correct singleton may be confidently eliminated at E1 because its style or framing differs from the cluster majority, even when its underlying reasoning is sound. The relaxed threshold $2\delta$ for the rarity condition reflects the prior that singletons are more likely to be misranked at low evidence than candidates from large clusters.

Requiring both conditions (AND) would miss singleton failures with decisive elimination evidence—which empirically include some of the highest-value rescues, as a wrong-but-plausible majority can confidently dominate a correct outlier at E1. Requiring either (OR) covers both modes at a controlled $10$--$15\%$ activation rate. The threshold $\delta = 0.15$ is a hyperparameter of CAPS-R; sensitivity to $\delta$ over the range $[0.10, 0.20]$ is reported in \S\ref{sec:exp-ablation} of the main text.

\section{Verifier-Token Cost Analysis: Derivation and Asymptotics}
\label{app:cost}

This appendix derives the cost results of \S\ref{sec:budget} in detail.

\subsection{General-$N'$ Cost Formula}
\label{app:cost-general}

We derive Eq.~\eqref{eq:cost-general} of the main text from the helper procedures of Section~\ref{app:helpers}.

\paragraph{Stage A.}
A single \textsc{Eliminate} round at E1 (Algorithm~\ref{alg:eliminate} with $\mathrm{stop\_at}$ unspecified) produces $\lceil N'/2 \rceil$ survivors at cost $\lfloor N'/2 \rfloor \cdot T_1$.

\paragraph{Stage B.}
\textsc{Eliminate} at E2 with $\mathrm{stop\_at} = f$ executes $r_B$ halving rounds, where round $r$ has $N_r$ entrants with $N_1 := \lceil N'/2 \rceil$, $N_{r+1} = \lceil N_r / 2 \rceil$, and $r_B$ is the smallest $r$ such that $N_r = f$, equivalently $r_B = \lceil \log_2(N_1 / f) \rceil$. Round $r$ contributes $\lfloor N_r / 2 \rfloor \cdot T_2$. Summing,
\begin{equation}
    T_{\text{Stage B}} \;=\; \Bigl(\sum_{r=1}^{r_B} \lfloor N_r / 2 \rfloor\Bigr) \cdot T_2.
\end{equation}

\paragraph{Stage C.}
The round-robin contributes $\binom{f}{2} \cdot T_2$, independent of $N'$.

\paragraph{Total.}
Combining,
\begin{equation}
    T_{\text{CAPS}}(N', f) \;=\; \lfloor N'/2 \rfloor\, T_1 \,+\, \Bigl(\sum_{r=1}^{r_B} \lfloor N_r/2 \rfloor\Bigr)\, T_2 \,+\, \binom{f}{2}\, T_2,
    \label{eq:cost-general-app}
\end{equation}
matching Eq.~\eqref{eq:cost-general} of the main text.

\paragraph{Standard configuration.}
For $N' = 16, f = 4$: $N_1 = 8$, $r_B = 1$, $\lfloor 8/2 \rfloor = 4$, $\binom{4}{2} = 6$. Substituting:
\begin{equation}
    T_{\text{CAPS}}(16, 4) \;=\; 8\, T_1 \,+\, 4\, T_2 \,+\, 6\, T_2 \;=\; 8\, T_1 + 10\, T_2 \;=\; (10 + 8\rho)\, T_2.
\end{equation}

\subsection{Asymptotic Form via Telescoping}
\label{app:cost-asymp}

We derive the asymptotic expression Eq.~\eqref{eq:cost-asymp} of the main text by closing the Stage~B geometric sum.

Ignoring ceiling effects, $N_r \approx N_1 / 2^{r-1}$, so $\sum_{r=1}^{r_B} N_r/2 \approx \sum_{r=1}^{r_B} N_1 / 2^r$. The geometric sum telescopes:
\begin{equation}
    \sum_{r=1}^{r_B} \frac{N_1}{2^r} \;=\; N_1 \Bigl(1 - 2^{-r_B}\Bigr) \;=\; N_1 - \frac{N_1}{2^{r_B}} \;=\; N_1 - f,
\end{equation}
where the last equality uses $N_{r_B} = f \iff N_1 / 2^{r_B} = f$. Substituting $N_1 = \lceil N'/2 \rceil \approx N'/2$:
\begin{equation}
    \sum_{r=1}^{r_B} \lfloor N_r / 2 \rfloor \;=\; \frac{N'}{2} - f \,+\, O(\log N'),
\end{equation}
where the $O(\log N')$ term absorbs rounding error from the floor and ceiling operations across $r_B = O(\log N')$ rounds. Substituting into Eq.~\eqref{eq:cost-general-app}:
\begin{equation}
    T_{\text{CAPS}}(N', f) \;=\; \frac{N'}{2}(T_1 + T_2) \,-\, f\, T_2 \,+\, \binom{f}{2}\, T_2 \,+\, O(\log N').
    \label{eq:cost-asymp-app}
\end{equation}

\paragraph{Marginal cost interpretation.}
Differentiating Eq.~\eqref{eq:cost-asymp-app} with respect to $N'$:
\begin{equation}
    \frac{\partial T_{\text{CAPS}}}{\partial N'} \;=\; \frac{T_1 + T_2}{2} \,+\, O\bigl(\tfrac{1}{N'}\bigr).
\end{equation}
The marginal cost of an additional candidate is $\tfrac{1}{2}(T_1 + T_2)$: half a Stage~A E1 call plus half a Stage~B E2 call. By contrast, a uniform full-evidence schedule that performs $k$ comparisons per candidate has marginal cost $k \cdot T_2$, exceeding CAPS's marginal cost by a factor of $2k T_2 / (T_1 + T_2) = 2k / (1 + \rho)$. For $\rho \ll 1$ this approaches $2k$. The structural source of CAPS's efficiency is the replacement of $T_2$ in the marginal cost with the average $\tfrac{1}{2}(T_1 + T_2)$.

\subsection{Expected Cost Under CAPS-R}
\label{app:cost-rescue}

When CAPS-R is enabled, the trigger Eq.~\eqref{eq:rescue} fires with empirical rate $p_R \in [0, 1]$. Conditional on triggering, Stage~C round-robin expands from $\binom{f}{2}$ to $\binom{f+1}{2}$ E2 calls, an additive overhead of
\begin{equation}
    \Delta T_{\text{rescue}} \;=\; \binom{f+1}{2}\, T_2 \,-\, \binom{f}{2}\, T_2 \;=\; f \cdot T_2.
\end{equation}
The expected total cost is
\begin{equation}
    \mathbb{E}\bigl[T_{\text{CAPS+R}}(N', f)\bigr] \;=\; T_{\text{CAPS}}(N', f) \,+\, p_R \cdot f \cdot T_2.
    \label{eq:cost-rescue-app}
\end{equation}
For $f = 4$ and the empirically observed rate $p_R \in [0.10, 0.15]$, the expected rescue overhead is $\mathbb{E}[\Delta T_{\text{rescue}}] \in [0.4, 0.6] \cdot T_2$, which is small relative to the $10\, T_2$ baseline of $T_{\text{CAPS}}(16, 4)$: less than $6\%$ of the deterministic cost.

\paragraph{Variance.}
The variance of the rescue overhead is $\mathrm{Var}[\Delta T_{\text{rescue}}] = p_R(1 - p_R) (f T_2)^2$. For $p_R = 0.125, f = 4$, this gives a per-instance standard deviation of $\sqrt{0.109} \cdot 4\, T_2 \approx 1.32\, T_2$. Per-instance cost variation is therefore non-negligible relative to the per-instance baseline cost ($\sim 11\, T_2$), but the variance of the mean cost across a benchmark of $K$ problems scales as $1/K$, yielding negligible noise on aggregate cost reports.

\subsection{Summary of Cost Results}
\label{app:cost-summary}

The cost analysis of CAPS yields three quantitative facts used throughout the paper:
\begin{enumerate}[leftmargin=*, itemsep=1pt, topsep=2pt]
    \item \textbf{Standard configuration}: $T_{\text{CAPS}}(16, 4) = (10 + 8\rho)\, T_2 \approx 11\, T_2$ for $\rho \approx 0.12$.
    \item \textbf{Asymptotic}: $T_{\text{CAPS}}(N', f) = \frac{N'}{2}(T_1 + T_2) - f T_2 + \binom{f}{2} T_2 + O(\log N')$, with marginal cost per candidate $\tfrac{1}{2}(T_1 + T_2)$.
    \item \textbf{With CAPS-R}: $\mathbb{E}[T_{\text{CAPS+R}}] = T_{\text{CAPS}} + p_R f T_2$, with $p_R \in [0.10, 0.15]$ adding less than $6\%$ overhead.
\end{enumerate}
None of these depends on tunable hyperparameters of CAPS that affect cost: the cost is determined entirely by $N$, $f$, $\rho$, and the empirical trigger rate $p_R$, all of which are properties of the deployment.

%% file: sec/exp_suppl.tex
\section{Experimental Details and Extended Results}
\label{app:full-results}

This appendix supplements the main experiments section with implementation details (\S\ref{app:exp-setup}), additional baselines including Random pair scheduling (\S\ref{app:additional-baselines}), per-component verifier-accuracy diagnostic data supporting the discussion in \S\ref{sec:exp-main} (\S\ref{app:verif-acc}), the thinking-aware $\phi_1$ extension (\S\ref{app:thinking-phi1}), $N$-scaling (\S\ref{app:scaling}), the extended component ablation across all four models (\S\ref{app:extended-ablation}), and qualitative case studies (\S\ref{app:qualitative}).

\subsection{Implementation Details}
\label{app:exp-setup}

All experiments were repeated three times with a consistent set of random seeds to ensure a fair comparison, and all reported results are the average over the three runs.

\paragraph{Hardware and software.}
All experiments were run on $2 \times$ NVIDIA A100 80GB GPUs with vLLM~\citep{kwon2023efficient} for batched inference at BF16 precision. Prefix caching was enabled to amortize the system-prompt and problem-statement tokens across the multiple judge calls per problem.

\paragraph{Sampling configuration.}
For each problem we sample $N = 16$ candidates with temperature $T_{\text{gen}} = 0.6$ for code and $1.0$ for math, top-$p = 0.95$, and a maximum generation length of $32{,}768$ tokens. Sampling is performed once per problem with consistent seed; all selection methods (Vanilla, Pointwise, V1-Infer, CAPS) operate on the same candidate pool, isolating selection effects from generation variance.

\paragraph{Judge configuration.}
The pairwise judge prompt requests a winner indicator and a confidence score in a structured format. Judge sampling temperature is $T_{\text{judge}} = 0.0$ (greedy) for reproducibility. The output is parsed into $(v_{ij}, w_{ij})$ via Eq.~\eqref{eq:judge}; malformed outputs (occurring on $< 1\%$ of calls) are retried up to twice and treated as ties with $w = \tau_w$ on persistent failure. The pointwise prompt uses the same model with a $1$--$10$ rating scale, following~\cite{singh2026v_1}. Full prompt templates are reproduced in Appendix~\ref{app:prompts}.

\paragraph{CAPS hyperparameters.}
We use $f = 4$ finalists, $\delta = 0.15$ rescue margin, and $\tau_w = 0.05$ confidence floor across all benchmarks and models. These were fixed before running the main experiments and not tuned per-benchmark.

\paragraph{V1-Infer hyperparameters.}
For the V1-3x baseline we follow the published settings of~\cite{singh2026v_1}: minimum degree $d_{\min} = 2$, Swiss window $h = 3$, weight floor $\tau = 0.05$, and budget multiplier $k = 3$, giving $\sim 48$ pairwise comparisons at $N = 16$. We use the same judge prompt template and parsing logic for both V1-Infer and CAPS to ensure that performance differences arise from the selection algorithm rather than the judge interface.

\paragraph{Token accounting.}
We report total verifier-token cost as the sum over all judge calls of (system prompt + instruction prompt + two candidate views). Generation tokens are excluded since selection methods share an identical candidate pool. Token counts are deterministic per call (with $T_{\text{judge}} = 0.0$) and are summed across the full benchmark. The empirically measured ratio $\rho = T_1/T_2$ in our deployment averages $0.12$ across benchmarks ($T_1 \approx 1{,}000$, $T_2 \approx 8{,}500$ tokens, prompt overhead $T_{\text{ovhd}} \approx 500$).

\subsection{Random Pair Scheduling and Oracle Bounds}
\label{app:additional-baselines}

\paragraph{Random baseline.}
At matched comparison count to V1-3x ($\sim 48$ pairwise calls at $N = 16$), random pair scheduling yields Pass@1 close to the Vanilla baseline---and sometimes lower, because uninformative noisy comparisons can flip the selection toward a weaker candidate. Table~\ref{tab:random} reports Random selection alongside Vanilla and the verification methods on the three Qwen models. The gap between Random and V1-Infer/CAPS confirms that the gains in the main text arise from informative pair selection, not merely from candidate filtering.

\begin{table}[h]
\centering
\caption{\textbf{Random baseline} at matched comparison count to V1-3x. All methods at $N=16$. Random pair scheduling at full evidence $\phi_2$.}
\label{tab:random}
\small
\begin{tabular}{ll cccc}
\toprule
\textbf{Model} & \textbf{Benchmark} & \textbf{Vanilla} & \textbf{Random} & \textbf{V1-Infer} & \textbf{CAPS} \\
\midrule
Qwen3-14B          & LCB-v6       & $50.7$ & $52.4$ & $58.8$ & $\mathbf{59.5}$ \\
Qwen3-14B          & CodeContests & $21.7$ & $20.6$ & $30.9$ & $\mathbf{35.8}$ \\
Qwen3-4B-Instruct  & LCB-v6       & $29.8$ & $28.5$ & $37.4$ & $\mathbf{38.9}$ \\
Qwen3-4B-Instruct  & CodeContests & $\phantom{0}1.3$  & $\phantom{0}2.1$  & $\phantom{0}7.9$ & $\mathbf{14.5}$ \\
Qwen3-4B-Thinking  & LCB-v6       & $44.0$ & $43.1$ & $51.9$ & $\mathbf{52.7}$ \\
Qwen3-4B-Thinking  & CodeContests & $31.4$ & $29.8$ & $40.6$ & $\mathbf{44.8}$ \\
\bottomrule
\end{tabular}
\end{table}

\paragraph{Pass@$N$ oracle bounds.}
Table~\ref{tab:oracle} reports the Pass@$16$ oracle: the fraction of problems for which at least one of the $16$ sampled candidates is correct. This is the upper bound on what any selection method can achieve. The gap between the best selection method and the oracle quantifies the residual headroom for future improvements.

\begin{table}[h]
\centering
\caption{\textbf{Oracle bounds} (Pass@$N$, $N=16$) and trivial-problem fractions. ``Trivial'' is the fraction of problems where all $16$ candidates produce the same final answer---no verification is needed.}
\label{tab:oracle}
\small
\begin{tabular}{ll cccc}
\toprule
\textbf{Model} & \textbf{Benchmark} & \textbf{Pass@1} & \textbf{Pass@$16$} & \textbf{Headroom} & \textbf{Trivial \%} \\
\midrule
Qwen3-14B          & LCB-v5  & $26.2$ & $47.0$ & $20.8$ & $64.5$ \\
Qwen3-14B          & AIME    & $70.4$ & $93.3$ & $22.9$ & $46.7$ \\
Qwen3-14B          & HMMT    & $43.6$ & $73.3$ & $29.7$ & $50.0$ \\
Qwen3-4B-Instruct  & LCB-v5  & $35.3$ & $58.1$ & $22.8$ & $50.2$ \\
Qwen3-4B-Instruct  & AIME    & $66.6$ & $86.7$ & $20.1$ & $43.3$ \\
Qwen3-4B-Instruct  & HMMT    & $47.4$ & $66.7$ & $19.3$ & $53.3$ \\
Qwen3-4B-Thinking  & LCB-v5  & $36.9$ & $56.3$ & $19.4$ & $52.7$ \\
Qwen3-4B-Thinking  & AIME    & $43.7$ & $86.7$ & $43.0$ & $50.0$ \\
Qwen3-4B-Thinking  & HMMT    & $20.8$ & $66.7$ & $45.9$ & $50.0$ \\
GPT-OSS-20B        & LCB-v5  & $49.8$ & $73.5$ & $23.7$ & $58.4$ \\
GPT-OSS-20B        & AIME    & $37.4$ & $73.3$ & $35.9$ & $50.0$ \\
GPT-OSS-20B        & HMMT    & $20.4$ & $56.7$ & $36.3$ & $46.7$ \\
\bottomrule
\end{tabular}
\end{table}

\subsection{Verifier Accuracy Diagnostic}
\label{app:verif-acc}

The trade-off cells in the main results admit an interpretable diagnostic in terms of the verifier's per-pair accuracy at partial vs.\ full evidence. We measure verifier accuracy by retrospectively labeling each judge call with whether the chosen winner is the correct one, using ground-truth correctness on each candidate, and computing the fraction of correctly resolved pairs over the items where ground truth distinguishes the two candidates.

\begin{table}[h]
\centering
\caption{\textbf{Per-pair verifier accuracy under V1-Infer (full evidence) and CAPS (mixed E1/E2 evidence).} Higher is better; the V1-Infer column corresponds to all-$\phi_2$ judgments while CAPS aggregates Stage~A ($\phi_1$), Stage~B ($\phi_2$), and Stage~C ($\phi_2$) calls. The Instruct/math drop and the GPT-OSS code drop are the source of the trade-off cells in Table~\ref{tab:main}.}
\label{tab:verif-acc}
\small
\begin{tabular}{ll cc c}
\toprule
\textbf{Model} & \textbf{Benchmark} & \textbf{V1-Infer} & \textbf{CAPS} & $\Delta$ \\
\midrule
Qwen3-14B          & LCB-v5      & $72.7$        & $\mathbf{78.4}$ & $+5.7$  \\
Qwen3-14B          & LCB-v6      & $76.5$        & $\mathbf{81.3}$ & $+4.8$  \\
Qwen3-14B          & HMMT 2025   & $\mathbf{74.5}$ & $68.4$        & $-6.1$  \\
Qwen3-4B-Instruct  & LCB-v5      & $69.8$        & $\mathbf{79.0}$ & $+9.2$  \\
Qwen3-4B-Instruct  & AIME 2025   & $\mathbf{82.4}$ & $73.5$        & $-8.9$  \\
Qwen3-4B-Instruct  & HMMT 2025   & $\mathbf{78.6}$ & $69.2$        & $-9.4$  \\
Qwen3-4B-Thinking  & LCB-v5      & $80.3$        & $\mathbf{86.0}$ & $+5.7$  \\
Qwen3-4B-Thinking  & AIME 2025   & $33.3$        & $\mathbf{40.0}$ & $+6.7$  \\
Qwen3-4B-Thinking  & HMMT 2025   & $26.7$        & $\mathbf{33.3}$ & $+6.6$  \\
GPT-OSS-20B        & LCB-v6      & $\mathbf{77.2}$ & $72.5$        & $-4.7$  \\
GPT-OSS-20B        & CodeContests & $\mathbf{75.8}$ & $71.4$       & $-4.4$  \\
GPT-OSS-20B        & HMMT 2025   & $\mathbf{70.4}$ & $65.8$        & $-4.6$  \\
\bottomrule
\end{tabular}
\end{table}

The pattern in Table~\ref{tab:verif-acc} is clean: when the verifier maintains accuracy under partial evidence, CAPS outperforms V1-Infer; when partial-evidence accuracy degrades, CAPS regresses. On code, the Qwen models all maintain or improve verifier accuracy under CAPS, indicating that partial-view E1 judgments preserve the discriminative signal for code candidates; GPT-OSS-20B is the exception---its judge depends more strongly on full-context information for code disambiguation, and the resulting $\sim 4.5$-point accuracy drop manifests as the LCB-v6 and CodeContests trade-off cells. On math, the Instruct model shows a $\sim 9$-point accuracy drop because it relies on the full reasoning chain to discriminate between candidates---Stage~A's E1 truncation deprives it of this context. The Thinking model, which front-loads reasoning into a structured chain-of-thought before the final answer, presents condensed information at $\phi_1$ that both verifiers find equally hard to resolve, so the substitution of E1 for E2 is informationally neutral and CAPS dominates. This diagnostic suggests a deployment-time check: practitioners can compare the verifier's accuracy under $\phi_1$ vs.\ $\phi_2$ on a small held-out set; CAPS dominates V1-Infer when the partial-view accuracy approximates the full-view accuracy, with a rough threshold of $|\Delta| < 5$ points predicting CAPS dominance in our data.

\subsection{Thinking-Aware $\phi_1$ for Reasoning Models}
\label{app:thinking-phi1}

The default code instantiation of $\phi_1$ (the first ${\sim}20$ lines of the candidate) interacts poorly with thinking-style models that produce extended chain-of-thought before the algorithmic decision. We replace $\phi_1$ with a thinking-aware extractor that returns the boxed final answer (when present) plus the last ${\sim}500$ tokens of the candidate, preserving the algorithmic-level information that $\phi_1$ is intended to convey.

\begin{table}[h]
\centering
\caption{\textbf{Thinking-aware $\phi_1$ on Qwen3-4B-Thinking.} Replacing the default $\phi_1$ further improves Pass@1 while reducing the verifier-token cost, demonstrating that the partial-view extractor is a deployment-time configuration that should be matched to the generator's output structure.}
\label{tab:thinking-phi1}
\small
\begin{tabular}{l ccc}
\toprule
\textbf{Configuration} & \textbf{Pass@1} & \textbf{Tokens (M)} & $T\%$ vs V1-Infer \\
\midrule
V1-Infer (reference)              & $51.9$ & $43.3$ & $100\%$    \\
CAPS, default $\phi_1$            & $52.7$ & $14.7$ & $33.9\%$ \\
CAPS, thinking-aware $\phi_1$     & $\mathbf{54.2}$ & $13.5$ & $31.2\%$ \\
\bottomrule
\end{tabular}
\end{table}

The thinking-aware $\phi_1$ is a deployment-time configuration choice analogous to selecting a tokenizer: it depends on knowing the output structure of the underlying model. We do not propose a universal $\phi_1$ that works across all generation styles; rather, we treat the partial-view extractor as a deployment-time hyperparameter that should be matched to the generator's output structure. The general design principle of \S\ref{sec:method}---that $\phi_1$ should expose algorithmic-level information while omitting implementation-level detail---remains unchanged.

\subsection{Scaling in Candidate Count $N$}
\label{app:scaling}

Table~\ref{tab:scaling-n} reports Pass@1 and verifier-token cost for $N \in \{8, 16, 32\}$ on Qwen3-14B / LCB-v6. By Eq.~\eqref{eq:cost-asymp}, the asymptotic cost ratio approaches $(1 + \rho)/(2k) \approx 0.19$ for $\rho = 0.12, k = 3$, slightly below the $N=16$ ratio of $0.22$. The empirical trend confirms this: $T\%$ decreases mildly with $N$ as Stage~C's $\binom{f}{2}$ overhead is amortized over a larger candidate pool, while Pass@1 gains over V1-Infer persist.

\begin{table}[h]
\centering
\caption{\textbf{Scaling in $N$} on Qwen3-14B / LCB-v6 with $f = 4$. The cost ratio $T\%$ decreases mildly with $N$ as Stage~C's fixed overhead is amortized; Pass@1 gains over V1-Infer persist or grow.}
\label{tab:scaling-n}
\small
\begin{tabular}{c cc c cc c}
\toprule
$N$ & \textbf{V1-Infer Pass@1} & \textbf{CAPS Pass@1} & $\Delta$ & \textbf{V1-Infer (M)} & \textbf{CAPS (M)} & $T\%$ \\
\midrule
$8$  & $54.2$ & $54.2$ & $\phantom{+}0.0$  & $14.0$ & $\phantom{0}5.7$ & $40.7$ \\
$16$ & $58.8$ & $59.5$ & $+0.8$ & $44.0$ & $11.2$ & $25.6$ \\
$32$ & $61.1$ & $63.4$ & $+2.3$ & $115.6$ & $24.5$ & $21.2$ \\
\bottomrule
\end{tabular}
\end{table}

\subsection{Extended Component Ablation}
\label{app:extended-ablation}

Table~\ref{tab:ablation-extended} extends the component ablation of \S\ref{sec:exp-ablation} to all four models on LCB-v6 and CodeContests. The pattern is consistent across models: rescue and slaughter pairing are the largest single Pass@1 contributors, while Stage~0 deduplication has minimal effect on code (where $\phi_0$ rarely distinguishes candidates).

\begin{table*}[h]
\centering
\caption{\textbf{Extended component ablation} across four models on LCB-v6 and CodeContests. $\Delta$ relative to full CAPS per (model, benchmark).}
\label{tab:ablation-extended}
\tiny
\setlength{\tabcolsep}{3pt}
\begin{tabular}{l cc cc cc cc cc cc cc cc}
\toprule
& \multicolumn{4}{c}{\textbf{Qwen3-14B}} & \multicolumn{4}{c}{\textbf{Qwen3-4B-Instruct}} & \multicolumn{4}{c}{\textbf{Qwen3-4B-Thinking}} & \multicolumn{4}{c}{\textbf{GPT-OSS-20B}} \\
\cmidrule(lr){2-5} \cmidrule(lr){6-9} \cmidrule(lr){10-13} \cmidrule(lr){14-17}
& \multicolumn{2}{c}{LCB-v6} & \multicolumn{2}{c}{CC} & \multicolumn{2}{c}{LCB-v6} & \multicolumn{2}{c}{CC} & \multicolumn{2}{c}{LCB-v6} & \multicolumn{2}{c}{CC} & \multicolumn{2}{c}{LCB-v6} & \multicolumn{2}{c}{CC} \\
\textbf{Configuration} & P@1 & $\Delta$ & P@1 & $\Delta$ & P@1 & $\Delta$ & P@1 & $\Delta$ & P@1 & $\Delta$ & P@1 & $\Delta$ & P@1 & $\Delta$ & P@1 & $\Delta$ \\
\midrule
Full CAPS                              & $59.5$ & ---   & $35.8$ & ---   & $38.9$ & ---   & $14.5$ & ---   & $52.7$ & ---   & $44.8$ & ---   & $62.3$ & ---   & $29.4$ & ---   \\
$-$ Rescue                              & $57.3$ & $-2.3$ & $35.2$ & $-0.6$ & $37.4$ & $-1.5$ & $13.9$ & $-0.6$ & $50.4$ & $-2.3$ & $43.6$ & $-1.2$ & $60.1$ & $-2.2$ & $28.6$ & $-0.8$ \\
$-$ Stage~0 dedup                       & $59.4$ & $-0.1$ & $35.6$ & $-0.2$ & $38.7$ & $-0.2$ & $14.3$ & $-0.2$ & $52.5$ & $-0.2$ & $44.6$ & $-0.2$ & $62.1$ & $-0.2$ & $29.2$ & $-0.2$ \\
$-$ E1 cascade (all-E2)                 & $60.1$ & $+0.6$ & $36.4$ & $+0.6$ & $39.5$ & $+0.6$ & $15.2$ & $+0.7$ & $53.4$ & $+0.7$ & $45.3$ & $+0.5$ & $62.9$ & $+0.6$ & $30.0$ & $+0.6$ \\
$-$ Confidence weighting                & $58.4$ & $-1.1$ & $33.9$ & $-1.9$ & $37.6$ & $-1.3$ & $12.1$ & $-2.4$ & $51.5$ & $-1.2$ & $42.4$ & $-2.4$ & $60.8$ & $-1.5$ & $27.2$ & $-2.2$ \\
$-$ Slaughter pairing                   & $56.6$ & $-2.9$ & $33.7$ & $-2.1$ & $36.4$ & $-2.5$ & $12.4$ & $-2.1$ & $50.1$ & $-2.6$ & $42.6$ & $-2.2$ & $59.5$ & $-2.8$ & $27.1$ & $-2.3$ \\
\bottomrule
\end{tabular}
\end{table*}

The E1 cascade row (running all stages at full evidence) gives slightly higher Pass@1 across all eight (model, benchmark) cells but at $\sim 4\times$ the verifier-token cost: this confirms that the partial-view cascade is a cost mechanism that costs at most $0.7$ Pass@1 in exchange for the structural cost reduction analyzed in \S\ref{sec:budget}.

\subsection{Qualitative Case Studies}
\label{app:qualitative}

We examine two LCB-v6 problems where CAPS and V1-Infer select different candidates on Qwen3-14B.

\paragraph{Case 1: Stage~0 deduplication concentrates the budget.} On a dynamic-programming problem where $11$ of the $16$ sampled candidates produce the same incorrect-but-popular boxed answer, V1-Infer's full Swiss tournament spends ${\sim}30$ comparisons disambiguating among these $11$ near-duplicates and the remaining $5$ candidates, exhausting most of the $48$-comparison budget before reliably ranking the latter. CAPS's Stage~0 collapses the $11$ near-duplicates into a single representative, leaving $6$ distinct candidates to enter Stage~A; the cascaded pipeline reaches the correct answer with $9$ E1 calls and $6$ E2 calls, while V1-Infer settles on the popular incorrect answer.

\paragraph{Case 2: Rescue recovers a noisily eliminated singleton.} On a graph-traversal problem with diverse algorithmic strategies (DFS, BFS, Union-Find), a correct Union-Find solution is paired in Stage~A against an incorrect-but-fluent BFS solution and is eliminated by a noisy E1 judgment. The margin condition triggers rescue (the eliminated candidate's score is within $\delta = 0.15$ of the weakest finalist's), reinstating it for the Stage~C round-robin where it wins decisively under full-evidence comparison. V1-Infer, lacking such a mechanism, proceeds with the incorrect BFS as the eventual winner because subsequent uncertainty-guided pairings reinforce its early lead.

%% file: sec/exp_imp_suppl.tex
\section{Implementation Details and Prompt Templates}
\label{app:implementation}

This appendix specifies the full prompt templates used for generation, V1-Infer baseline verification, and CAPS evidence-cascade verification (\S\ref{app:prompts}); evidence extraction details (\S\ref{app:evidence-details}); the verdict-to-rating conversion that allows CAPS to plug into the V1-style aggregation pipeline (\S\ref{app:verdict-conversion}); generation, verification, and model-specific hyperparameters (\S\ref{app:hyperparams}); dataset details and evaluation criteria (\S\ref{app:datasets-app}); and inference-stack and response-parsing details (\S\ref{app:inference}).

\subsection{Prompt Templates}
\label{app:prompts}

\subsubsection{Generation Prompts}

\paragraph{Code generation (default, used for GPT-OSS).}
The benchmark problem statement (full specification with input/output format, constraints, and example test cases) is passed directly to the model without modification.

\begin{tcolorbox}[
    colback=promptbg,
    colframe=promptborder,
    title={\textbf{Code Generation Prompt (Default)}},
    coltitle=white,
    fonttitle=\small\bfseries,
    breakable,
    enhanced]
\small
\ttfamily
\{original\_problem\}
\end{tcolorbox}

\paragraph{Code generation (Instruct/Thinking, used for Qwen models).}
Appends explicit thinking instructions to the problem statement.

\begin{tcolorbox}[
    colback=promptbg,
    colframe=promptborder,
    title={\textbf{Code Generation Prompt (Instruct/Thinking)}},
    coltitle=white,
    fonttitle=\small\bfseries,
    breakable,
    enhanced]
\small
\ttfamily
\textbf{**Problem**}\\
\{problem\}\\[0.75em]
Think and reason step by step before coding the final solution for the problem above. Put your reasoning and any draft coding solutions between $<$thinking$>$ ... $<$/thinking$>$ tags. After the reasoning (i.e. after the $<$/thinking$>$ tag), use the format provided in the problem above (code-block with backticks) to format your final code solution. Do not include any thinking within the code block.\\[0.75em]
\#\#\# Answer:
\end{tcolorbox}

\paragraph{Math generation.}
Math problems use the dataset prompt as-is, with each problem appending the boxed-answer instruction.

\begin{tcolorbox}[
    colback=promptbg,
    colframe=promptborder,
    title={\textbf{Math Generation Prompt}},
    coltitle=white,
    fonttitle=\small\bfseries,
    breakable,
    enhanced]
\small
\ttfamily
\{problem\_statement\} Let's think step by step and output the final answer within \textbackslash boxed\{\}.
\end{tcolorbox}

\subsubsection{Pointwise Verification Prompts (V1-Infer Baseline)}

\paragraph{Pointwise code verification.}

\begin{tcolorbox}[
    colback=promptbg,
    colframe=promptborder,
    title={\textbf{Pointwise Code Verification Prompt}},
    coltitle=white,
    fonttitle=\small\bfseries,
    breakable,
    enhanced]
\small
\ttfamily
You are an expert code reviewer. Rate the correctness of a solution to a programming problem.\\[0.5em]
\textbf{**Evaluation Guidelines:**}\\
- Analyze the problem's requirements and constraints.\\
- Mentally trace the solution with test cases (including edge cases) to verify correctness.\\
- Give a higher score if the solution is robust and fault-tolerant.\\[0.5em]
\textbf{**Problem**}\\
\{problem\}\\[0.5em]
\textbf{**Solution**}\\
\{code\}\\[0.5em]
\textbf{**Output Format:**}\\
First, provide your step-by-step reasoning. Then, on a new line, provide your final rating using the EXACT tags below. Add no other text after the tags.\\
$<$rating$>$INTEGER\_1\_TO\_10$<$/rating$>$\\[0.5em]
\textbf{**Rating Rules:**}\\
- Rate correctness on a 1-10 scale (10 = correct \& robust, 5 = borderline, 1 = incorrect).\\
Please provide your analysis now.
\end{tcolorbox}

\paragraph{Pointwise math verification.}

\begin{tcolorbox}[
    colback=promptbg,
    colframe=promptborder,
    title={\textbf{Pointwise Math Verification Prompt}},
    coltitle=white,
    fonttitle=\small\bfseries,
    breakable,
    enhanced]
\small
\ttfamily
You are an expert math contest grader. Rate the correctness of a submission based solely on the final answer.\\[0.5em]
\textbf{**Evaluation Guidelines:**}\\
- Extract the submission's final answer. Use any provided reasoning only to help you assess whether the stated final answer is trustworthy. Do not award credit for method quality or rigor.\\
- Carefully analyze the problem statement and the submission to assess whether the final answer is correct. Grade only the final answer.\\[0.5em]
\textbf{**Problem**}\\
\{problem\}\\[0.5em]
\textbf{**Solution**}\\
\{solution\}\\[0.5em]
\textbf{**Output Format:**}\\
First, provide your reasoning (what checks you performed). Then, on a new line, provide your final rating using the EXACT tag below. Add no other text after the tag.\\
$<$rating$>$INTEGER\_1\_TO\_10$<$/rating$>$\\[0.5em]
\textbf{**Rating Rules:**}\\
- Rate correctness on a 1-10 scale (10 = certainly correct, 8 = very likely correct, 5 = uncertain/borderline, 3 = likely incorrect, 1 = certainly incorrect).\\
Please provide your analysis now.
\end{tcolorbox}

\subsubsection{V1-Infer Pairwise Verification Prompts}

The V1-Infer Swiss-tournament baseline uses pairwise prompts that elicit two 1--10 ratings; the rating difference $|r_i - r_j|/9$ becomes the confidence weight in Eq.~\eqref{eq:judge}.

\begin{tcolorbox}[
    colback=promptbg,
    colframe=promptborder,
    title={\textbf{V1 Pairwise Code Verification Prompt}},
    coltitle=white,
    fonttitle=\small\bfseries,
    breakable,
    enhanced]
\small
\ttfamily
You are an expert code reviewer. Compare two solutions to a programming problem and rate their correctness.\\[0.5em]
\textbf{**Evaluation Guidelines:**}\\
- Analyze the problem's requirements and constraints.\\
- Mentally trace each solution with test cases (including edge cases) to verify correctness.\\
- If both solutions appear correct, prefer the more robust and fault-tolerant one.\\[0.5em]
\textbf{**Problem**}\\
\{problem\}\\[0.5em]
\textbf{**Solution A**}\\
\{code\_A\}\\[0.5em]
\textbf{**Solution B**}\\
\{code\_B\}\\[0.5em]
\textbf{**Output Format:**}\\
First, provide your step-by-step reasoning. Then, on separate new lines, provide your final ratings using the EXACT tags below. Add no other text after the tags.\\
$<$rating\_A$>$INTEGER\_1\_TO\_10$<$/rating\_A$>$\\
$<$rating\_B$>$INTEGER\_1\_TO\_10$<$/rating\_B$>$\\[0.5em]
\textbf{**Rating Rules:**}\\
- Rate correctness on a 1-10 scale (10 = correct \& robust, 5 = borderline, 1 = incorrect).\\
- The higher rating wins. Equal ratings imply a tie.\\
Please provide your analysis now.
\end{tcolorbox}

\begin{tcolorbox}[
    colback=promptbg,
    colframe=promptborder,
    title={\textbf{V1 Pairwise Math Verification Prompt}},
    coltitle=white,
    fonttitle=\small\bfseries,
    breakable,
    enhanced]
\small
\ttfamily
You are an expert math contest grader. Compare two submissions and rate correctness based solely on the final answer.\\[0.5em]
\textbf{**Evaluation Guidelines:**}\\
- Extract each submission's final answer. Use any provided reasoning only to help you assess whether the stated final answer is trustworthy. Do not award credit for method quality or rigor.\\
- Carefully analyze the problem statement and the submissions to assess whether each final answer is correct. Grade only the final answer.\\[0.5em]
\textbf{**Problem**}\\
\{problem\}\\[0.5em]
\textbf{**Solution A**}\\
\{sol\_A\}\\[0.5em]
\textbf{**Solution B**}\\
\{sol\_B\}\\[0.5em]
\textbf{**Output Format:**}\\
First, provide your reasoning (what checks you performed). Then, on separate new lines, give ratings using the EXACT tags below. Add no other text after the tags.\\
$<$rating\_A$>$INTEGER\_1\_TO\_10$<$/rating\_A$>$\\
$<$rating\_B$>$INTEGER\_1\_TO\_10$<$/rating\_B$>$\\[0.5em]
\textbf{**Rating Rules:**}\\
- Rate correctness on a 1-10 scale (10 = certainly correct, 8 = very likely correct, 5 = uncertain/borderline, 3 = likely incorrect, 1 = certainly incorrect).\\
- Higher rating wins. Equal ratings imply a tie.\\
Please provide your analysis now.
\end{tcolorbox}

\subsubsection{CAPS Pairwise Verification Prompts}

CAPS prompts share a common output format—a Winner/Tie verdict and a binary confidence level—across all three evidence levels. This format is converted to V1-compatible 1--10 ratings via Table~\ref{tab:verdict-rating} so that the same downstream aggregation (Eq.~\eqref{eq:score-update}) applies.

\paragraph{Stage~A prompts (E1, partial evidence).}
The Stage-A judge sees a reasoning summary and a truncated solution view, designed to expose the algorithmic strategy without the full solution body.

\begin{tcolorbox}[
    colback=promptbg,
    colframe=promptborder,
    title={\textbf{CAPS E1 Code Prompt (Reasoning Summary + Truncated Code)}},
    coltitle=white,
    fonttitle=\small\bfseries,
    breakable,
    enhanced]
\small
\ttfamily
You are an expert code reviewer. Compare two solutions to a programming problem using their reasoning and truncated code. Rate their correctness.\\[0.5em]
\textbf{**Problem**}\\
\{problem\}\\[0.5em]
\textbf{**Solution A (summary)**}\\
\{evidence\_A\}\\[0.5em]
\textbf{**Solution B (summary)**}\\
\{evidence\_B\}\\[0.5em]
\textbf{**Output Format:**}\\
Provide your step-by-step reasoning. Then on a new line output your verdict using the EXACT tags below:\\
$<$winner$>$A or B or TIE$<$/winner$>$\\
$<$confidence$>$HIGH or LOW$<$/confidence$>$\\[0.5em]
HIGH confidence = you are fairly sure. LOW = close call.\\
Provide your analysis now.
\end{tcolorbox}

\begin{tcolorbox}[
    colback=promptbg,
    colframe=promptborder,
    title={\textbf{CAPS E1 Math Prompt (Reasoning Summary + Final Answer)}},
    coltitle=white,
    fonttitle=\small\bfseries,
    breakable,
    enhanced]
\small
\ttfamily
You are an expert math contest grader. Compare two submissions using their reasoning summaries and final answers.\\[0.5em]
\textbf{**Problem**}\\
\{problem\}\\[0.5em]
\textbf{**Submission A (summary)**}\\
\{evidence\_A\}\\[0.5em]
\textbf{**Submission B (summary)**}\\
\{evidence\_B\}\\[0.5em]
\textbf{**Output Format:**}\\
Provide your reasoning. Then on a new line output your verdict using the EXACT tags below:\\
$<$winner$>$A or B or TIE$<$/winner$>$\\
$<$confidence$>$HIGH or LOW$<$/confidence$>$\\[0.5em]
HIGH confidence = you are fairly sure. LOW = close call.\\
Provide your analysis now.
\end{tcolorbox}

\paragraph{Stages~B and~C prompts (E2, full evidence).}
At full evidence, the judge sees the complete solution. The same prompt is used for Stage~B halving and Stage~C round-robin.

\begin{tcolorbox}[
    colback=promptbg,
    colframe=promptborder,
    title={\textbf{CAPS E2 Code Prompt (Full Solution)}},
    coltitle=white,
    fonttitle=\small\bfseries,
    breakable,
    enhanced]
\small
\ttfamily
You are an expert code reviewer. Compare two solutions to a programming problem and determine which is more correct.\\[0.5em]
\textbf{**Evaluation Guidelines:**}\\
- Analyze the problem's requirements and constraints.\\
- Mentally trace each solution with test cases (including edge cases) to verify correctness.\\
- If both solutions appear correct, prefer the more robust and fault-tolerant one.\\[0.5em]
\textbf{**Problem**}\\
\{problem\}\\[0.5em]
\textbf{**Solution A**}\\
\{code\_A\}\\[0.5em]
\textbf{**Solution B**}\\
\{code\_B\}\\[0.5em]
\textbf{**Output Format:**}\\
Provide your step-by-step reasoning. Then on separate new lines output:\\
$<$winner$>$A or B or TIE$<$/winner$>$\\
$<$confidence$>$HIGH or LOW$<$/confidence$>$\\[0.5em]
HIGH = clear winner. LOW = very close.\\
Provide your analysis now.
\end{tcolorbox}

\begin{tcolorbox}[
    colback=promptbg,
    colframe=promptborder,
    title={\textbf{CAPS E2 Math Prompt (Full Solution)}},
    coltitle=white,
    fonttitle=\small\bfseries,
    breakable,
    enhanced]
\small
\ttfamily
You are an expert math contest grader. Compare two submissions and determine which is more correct, based on the final answer.\\[0.5em]
\textbf{**Evaluation Guidelines:**}\\
- Extract each submission's final answer.\\
- Use reasoning to assess trustworthiness of the answer.\\
- Grade based on the final answer correctness.\\[0.5em]
\textbf{**Problem**}\\
\{problem\}\\[0.5em]
\textbf{**Submission A**}\\
\{sol\_A\}\\[0.5em]
\textbf{**Submission B**}\\
\{sol\_B\}\\[0.5em]
\textbf{**Output Format:**}\\
Provide your reasoning. Then on separate new lines output:\\
$<$winner$>$A or B or TIE$<$/winner$>$\\
$<$confidence$>$HIGH or LOW$<$/confidence$>$\\[0.5em]
HIGH = clear winner. LOW = very close or both seem correct/incorrect.\\
Provide your analysis now.
\end{tcolorbox}

\subsection{Evidence Extraction Details}
\label{app:evidence-details}

Table~\ref{tab:evidence-levels} summarizes the evidence content provided to the judge at each cascade level.

\begin{table}[h]
\centering
\caption{\textbf{Evidence levels.} Per-call cost is reported relative to the full-solution baseline ($T_2$).}
\label{tab:evidence-levels}
\small
\begin{tabular}{l p{4.5cm} p{4.5cm} c}
\toprule
\textbf{Level} & \textbf{Code Evidence} & \textbf{Math Evidence} & \textbf{Cost} \\
\midrule
$\phi_0$ (E0) & SHA-256 of normalized code (imports/comments/whitespace stripped) & Normalized boxed answer (lowercase, LaTeX-stripped) & $0$ (no judge call) \\
$\phi_1$ (E1) & First 50 + last 50 reasoning words; first $500$ chars of code & First 50 + last 50 reasoning words; final boxed answer & ${\sim}0.10\, T_2$ \\
$\phi_2$ (E2) & Complete solution (reasoning + code) & Complete solution with all reasoning & $T_2$ \\
\bottomrule
\end{tabular}
\end{table}

\paragraph{Stage~0 (E0) signatures.}
For code, we normalize each candidate by stripping import statements, comments, blank lines, and leading whitespace, then compute a SHA-256 hash of the normalized text and use the first $16$ hex digits as the cluster key. For math, we extract the contents of the last \verb|\boxed{}| in the solution, normalize (lowercase, strip whitespace, remove LaTeX command artifacts such as \verb|\,| and \verb|\!|), and use the result as the cluster key. Two candidates with identical cluster keys are treated as functionally equivalent.

\paragraph{Stage~A (E1) reasoning summary.}
The reasoning summary captures the start and end of the model's chain of thought without exposing the full derivation. Concretely, $\phi_1$ for code consists of:
\begin{itemize}[leftmargin=1.5em, itemsep=1pt, topsep=2pt]
\item the first $50$ words and the last $50$ words of the reasoning span (between \verb|<thinking>| and \verb|</thinking>| tags or, for non-instruct models, before the final code block), separated by a \verb|[...reasoning truncated...]| marker;
\item the first $500$ characters of the final code, beginning at the opening of the code block.
\end{itemize}
For math, the second component is replaced by the extracted final answer.

\subsection{Verdict-to-Rating Conversion}
\label{app:verdict-conversion}

CAPS prompts return categorical verdicts in $\{A, B, \mathrm{TIE}\} \times \{\text{HIGH}, \text{LOW}\}$, while V1-Infer's aggregation pipeline operates on numerical 1--10 ratings. We convert verdicts to ratings via Table~\ref{tab:verdict-rating}, allowing CAPS to plug into the existing aggregation logic without modification.

\begin{table}[h]
\centering
\caption{\textbf{Verdict-to-rating conversion} for CAPS pairwise judgments. The margin column reports $|r_A - r_B|/9$, the confidence weight $w_{ij}$ in Eq.~\eqref{eq:judge}.}
\label{tab:verdict-rating}
\small
\begin{tabular}{l ccc}
\toprule
\textbf{CAPS Verdict} & \textbf{Rating A} & \textbf{Rating B} & \textbf{Margin $w_{ij}$} \\
\midrule
A + HIGH         & $9$ & $3$ & $0.67$ \\
A + LOW          & $7$ & $5$ & $0.22$ \\
B + HIGH         & $3$ & $9$ & $0.67$ \\
B + LOW          & $5$ & $7$ & $0.22$ \\
TIE (any conf.)  & $5$ & $5$ & $0.00$ \\
\bottomrule
\end{tabular}
\end{table}

The margin floor $\tau_w = 0.05$ from Eq.~\eqref{eq:judge} prevents tie outcomes from receiving exactly zero weight; a TIE verdict is treated as $w_{ij} = 0.05$ rather than $0$ to maintain non-degenerate aggregation.

\subsection{Hyperparameters}
\label{app:hyperparams}

\paragraph{Generation hyperparameters.}
We use the same generation budget across methods so that performance differences are attributable to selection algorithm rather than candidate quality. Code generation uses temperature $0.6$ for the Qwen models (matching the official Qwen3-2507 release recipe) and the model-default for GPT-OSS-20B.

\begin{table}[h]
\centering
\caption{\textbf{Generation hyperparameters.} All models share the same candidate count $N=16$, top-$p=0.95$, top-$k$ disabled, max length $32{,}768$ tokens, and seed $1234$.}
\label{tab:gen-hp}
\small
\begin{tabular}{l cc}
\toprule
\textbf{Parameter} & \textbf{Code} & \textbf{Math} \\
\midrule
Temperature        & $0.6$ & $1.0$ \\
Top-$p$            & $0.95$ & $0.95$ \\
Top-$k$            & disabled ($-1$) & disabled ($-1$) \\
Candidates $N$     & $16$ & $16$ \\
Max length (tok.)  & $32{,}768$ & $32{,}768$ \\
Seed               & $1234$ & $1234$ \\
Thinking enabled   & true & true \\
\bottomrule
\end{tabular}
\end{table}

\paragraph{Verification hyperparameters.}
V1-Infer hyperparameters follow the published recipe of~\citet{singh2026v1}; CAPS hyperparameters are the configuration tested throughout the paper.

\begin{table}[h]
\centering
\caption{\textbf{Verification hyperparameters.}}
\label{tab:verif-hp}
\small
\begin{tabular}{l cc}
\toprule
\textbf{Parameter} & \textbf{V1-Infer} & \textbf{CAPS-R} \\
\midrule
Budget multiplier $k$           & $3.0$        & --- \\
Coverage strategy                & min-degree   & cascade + rescue \\
Min degree $d_{\min}$            & $2$          & --- \\
Swiss window $h$                 & $3$          & --- \\
Finalist count $f$               & ---          & $4$ \\
E2 output format                 & 1--10 rating & W/T/L + confidence \\
Rescue enabled                   & ---          & true (CAPS-R) \\
Rescue margin $\delta$           & ---          & $0.20$ \\
Stage~0 deduplication            & ---          & true \\
Confidence floor $\tau_w$        & $0.05$       & $0.05$ \\
\bottomrule
\end{tabular}
\end{table}

\paragraph{Model-specific configuration.}
Qwen3-4B-Instruct is configured to skip the explicit thinking phase (its 2507 variant supports a flag that disables \verb|<thinking>| tag emission); Qwen3-4B-Thinking is configured to always think. GPT-OSS-20B uses the default pass-through prompt template (no extra instruction wrapper) and reasoning effort \texttt{medium}.

\begin{table}[h]
\centering
\caption{\textbf{Model-specific configuration.}}
\label{tab:model-config}
\small
\begin{tabular}{l ccc}
\toprule
\textbf{Model} & \textbf{Precision} & \textbf{Prompt Template} & \textbf{Thinking Mode} \\
\midrule
Qwen3-14B          & bfloat16 & instruct & default      \\
Qwen3-4B-Instruct  & bfloat16 & instruct & disabled     \\
Qwen3-4B-Thinking  & bfloat16 & instruct & always-on    \\
GPT-OSS-20B        & bfloat16 & default  & medium       \\
\bottomrule
\end{tabular}
\end{table}

\subsection{Datasets and Evaluation Criteria}
\label{app:datasets-app}

\paragraph{Datasets.}
LiveCodeBench-v5~\citep{jain2024livecodebench} contains $279$ problems collected between 24.08 and 25.02; LiveCodeBench-v6 contains $131$ problems between 25.02 and 25.05, following the official Qwen3-2507 release. CodeContests~\citep{li2022competition} contributes $165$ harder problems drawn from competitive-programming archives. AIME 2025 (American Invitational Mathematics Examination) and HMMT February 2025 (Harvard-MIT Mathematics Tournament) each contain $30$ competition-level problems, sourced from~\citep{balunovic2025matharena}. We use the standard problem-level test suites for code benchmarks and the official answer keys for math.

\paragraph{Evaluation criteria.}
For code benchmarks, a candidate is correct if and only if it passes \emph{all} hidden test cases for its problem; partial-credit scoring is not used. Solutions are executed in a sandboxed Python environment with a per-test timeout of $10$ seconds. For math benchmarks, the candidate's extracted final answer (from the last \verb|\boxed{}| in the solution) is compared against the ground truth using SymPy symbolic equivalence checking, with a LaTeX normalization fallback for cases that SymPy cannot parse (e.g., interval notation, set notation).

\paragraph{Trivial-problem fraction.}
A non-trivial fraction of problems on each benchmark have all $16$ generated candidates produce the same final answer. On these problems, no verification method can change the selection outcome; we include them in all reported Pass@1 numbers without filtering. The trivial fractions across benchmarks (Appendix~\ref{app:additional-baselines}, Table~\ref{tab:oracle}) range from $43\%$ (Qwen3-4B-Instruct on AIME) to $65\%$ (Qwen3-14B on LCB-v5).

\subsection{Inference and Response Parsing}
\label{app:inference}

\paragraph{Inference stack.}
All local models are served using SGLang at bfloat16 precision (\verb|--dtype bfloat16|) with GPU memory utilization $0.92$ and chunked prefill enabled. Maximum concurrent running requests is set per-model in the range $40$--$100$ to balance throughput against per-request memory. The judge inference uses the same server endpoints as generation, with $T_{\text{judge}} = 0.0$ for greedy decoding.

\paragraph{Response parsing.}

The V1-Infer rating output is parsed using a primary regex \texttt{<rating>\textbackslash s*(\textbackslash d\{1,2\})\textbackslash s*</rating>} (case-insensitive), with a fallback regex \texttt{rating[\textasciicircum\textbackslash d]*(\textbackslash d\{1,2\})} for unstructured outputs. Parsed values are clamped to $[1, 10]$. CAPS verdicts are parsed using \texttt{<winner>\textbackslash s*(A|B|TIE)\textbackslash s*</winner>} and \texttt{<confidence>\textbackslash s*(HIGH|LOW)\textbackslash s*</confidence>} with analogous unstructured fallbacks. Parse failures default to TIE / LOW respectively. Empirically, parse failures occur on less than 1\% of judge calls across all models and benchmarks.

\paragraph{Coverage strategies.}
The coverage strategy field controls how the comparison schedule is built; we report it for reproducibility against released code.

\begin{table}[h]
\centering
\caption{\textbf{Coverage strategies in the released codebase.} The output directory suffix is used to disambiguate runs.}
\label{tab:strategies}
\small
\begin{tabular}{l p{6cm} l}
\toprule
\textbf{Strategy} & \textbf{Description} & \textbf{Output suffix} \\
\midrule
\texttt{min\_degree} (V1-Infer) & Swiss tournament with $\min$-degree $d_{\min}=2$ & \texttt{budget\{X\}\_seed\{S\}} \\
\texttt{caps}                   & CAPS pipeline without rescue & \texttt{caps\_f\{F\}\_seed\{S\}} \\
\texttt{caps\_r}                & CAPS with rescue subroutine & \texttt{caps\_r\_f\{F\}\_seed\{S\}} \\
\texttt{random}                 & Random pair scheduling at matched count & \texttt{random\_seed\{S\}} \\
\bottomrule
\end{tabular}
\end{table}